\documentclass[10pt,journal,compsoc]{IEEEtran}

\usepackage[english]{babel}
\usepackage{graphicx}
\usepackage{amsmath}
\usepackage{array}
\usepackage{algorithm}
\usepackage{algorithmic}
\usepackage{multirow}
\usepackage{color, colortbl}

\addto\captionsenglish{\renewcommand{\figurename}{Fig.}}
\addto\captionsenglish{\renewcommand{\tablename}{TABLE}}
\DeclareMathOperator*{\argmin}{arg\,min}
\DeclareMathOperator*{\intersec}{intersect}
\newcommand{\tabincell}[2]{\begin{tabular}{@{}#1@{}}#2\end{tabular}}

\usepackage{expl3}
\ExplSyntaxOn
\newcommand\latinabbrev[1]{
  \peek_meaning:NTF . {% Same as \@ifnextchar
    #1\@}%
  { \peek_catcode:NTF a {% Check whether next char has same catcode as \'a, i.e., is a letter
      #1.\@ }%
    {#1.\@}}}
\ExplSyntaxOff

\def\eg{\latinabbrev{e.g}}
\def\etal{\latinabbrev{et al}}

\def\ie{\latinabbrev{i.e}}

\ifCLASSOPTIONcompsoc
  \usepackage[nocompress]{cite}
\else
  \usepackage{cite}
\fi
 
\ifCLASSOPTIONcompsoc
 \usepackage[caption=false,font=footnotesize,labelfont=sf,textfont=sf]{subfig}
\else
 \usepackage[caption=false,font=footnotesize]{subfig}
\fi

\usepackage{fixltx2e}

\begin{document}

\title{Visual and Semantic Knowledge Transfer for Large Scale Semi-supervised Object Detection}

\author{Yuxing~Tang,
        Josiah~Wang,
        Xiaofang~Wang,
        Boyang~Gao, 
        Emmanuel~Dellandr\'ea,
        Robert~Gaizauskas
        and~Liming~Chen~\IEEEmembership{Senior Member, ~IEEE}% <-this % stops a space
\IEEEcompsocitemizethanks{
\IEEEcompsocthanksitem Y. Tang is with Imaging Biomarkers and Computer-Aided Diagnosis Laboratory, National Institutes of Health (NIH) Clinical Center, 10 Center Dr., Bethesda, 20814, USA. This work was done when he was a Ph.D. student in LIRIS, \'Ecole Centrale de Lyon, France.\protect\\
E-mail: yuxing.tang@ec-lyon.fr
\IEEEcompsocthanksitem X. Wang, E. Dellandr\'ea and L. Chen are with LIRIS, CNRS UMR 5205, \'Ecole Centrale de Lyon, 36 avenue Guy de Collongue, \'Ecully, F-69134, France.\protect\\
E-mail: \{xiaofang.wang, emmanuel.dellandrea, liming.chen\}@ec-lyon.fr
\IEEEcompsocthanksitem J. Wang and R. Gaizauskas are with Department of Computer Science, The University of Sheffield, Regent Court, 211 Portobello Street, Sheffield S1 4DP, United Kingdom. \protect\\
E-mail: \{j.k.wang, r.gaizauskas\}@sheffield.ac.uk
\IEEEcompsocthanksitem B. Gao is with Department of Advanced Robotics (ADVR), Istituto Italiano di Tecnologia (IIT), Via Morego, Genova, 16163, Italy. \protect\\
E-mail: boyang.gao@iit.it
}% <-this % stops an unwanted space
% \thanks{Manuscript received April 19, 2005; revised August 26, 2015.}
}

%{Shell \MakeLowercase{\textit{et al.}}: Bare Demo of IEEEtran.cls for Computer Society Journals}

\IEEEtitleabstractindextext{%
\begin{abstract}
Deep CNN-based object detection systems have achieved remarkable success on several large-scale object detection benchmarks. However, training such detectors requires a large number of labeled bounding boxes, which are more difficult to obtain than image-level annotations. Previous work addresses this issue by transforming image-level classifiers into object detectors. This is done by modeling the differences between the two on categories with both image-level and bounding box annotations, and transferring this information to convert classifiers to detectors for categories without bounding box annotations. We improve this previous work by incorporating knowledge about object similarities from visual and semantic domains during the transfer process. The intuition behind our proposed method is that visually and semantically similar categories should exhibit more common transferable properties than dissimilar categories, e.g.\ a better detector would result by transforming the differences between a dog classifier and a dog detector onto the cat class, than would by transforming from the violin class. Experimental results on the challenging ILSVRC2013 detection dataset demonstrate that each of our proposed object similarity based knowledge transfer methods outperforms the baseline methods. We found strong evidence that visual similarity and semantic relatedness are complementary for the task, and when combined notably improve detection, achieving state-of-the-art detection performance in a semi-supervised setting.
\end{abstract}

}

\maketitle

\IEEEdisplaynontitleabstractindextext

\IEEEpeerreviewmaketitle

\IEEEraisesectionheading{\section{Introduction}\label{sec:introduction}}

\IEEEPARstart{G}{iven} an image, an object detection/localization method aims to recognize and locate objects of interest within it. It is one of the most widely studied problems in computer vision with a variety of applications. Most object detectors adopt strong supervision in learning appearance models of object categories, that is by using training images annotated with bounding boxes encompassing the objects of interest, along with their category labels. 
The recent success of deep convolutional neural networks (CNN) \cite{AlexNet_NIPS2012}
for object detection, such as DetectorNet \cite{Szegedy_NIPS2013}, OverFeat
\cite{sermanet-iclr-14}, R-CNN \cite{girshick2014rcnn}, SPP-net \cite{SPPnet_TPAMI2015},
Fast R-CNN \cite{girshick15fastrcnn}, Faster R-CNN \cite{ren15fasterrcnn},
YOLO \cite{Redmon_2016_CVPR} and SSD \cite{liu2016single}, is heavily dependent on a large amount of training data manually labeled with object localizations (\eg, PASCAL VOC \cite{Everingham-et-al-IJCV10}, ILSVRC (subset of ImageNet) \cite{ILSVRC15}, and Microsoft COCO \cite{MSCOCO14} datasets).

%%%%%%%%%%%%%%%%Fig intro%%%%%%%%%%%%%%%%%%%%%%%%%%%%%%%%%%%%%%%%%%
\begin{figure*}[htbp!]
  \centering
  \includegraphics[width=0.95\linewidth]{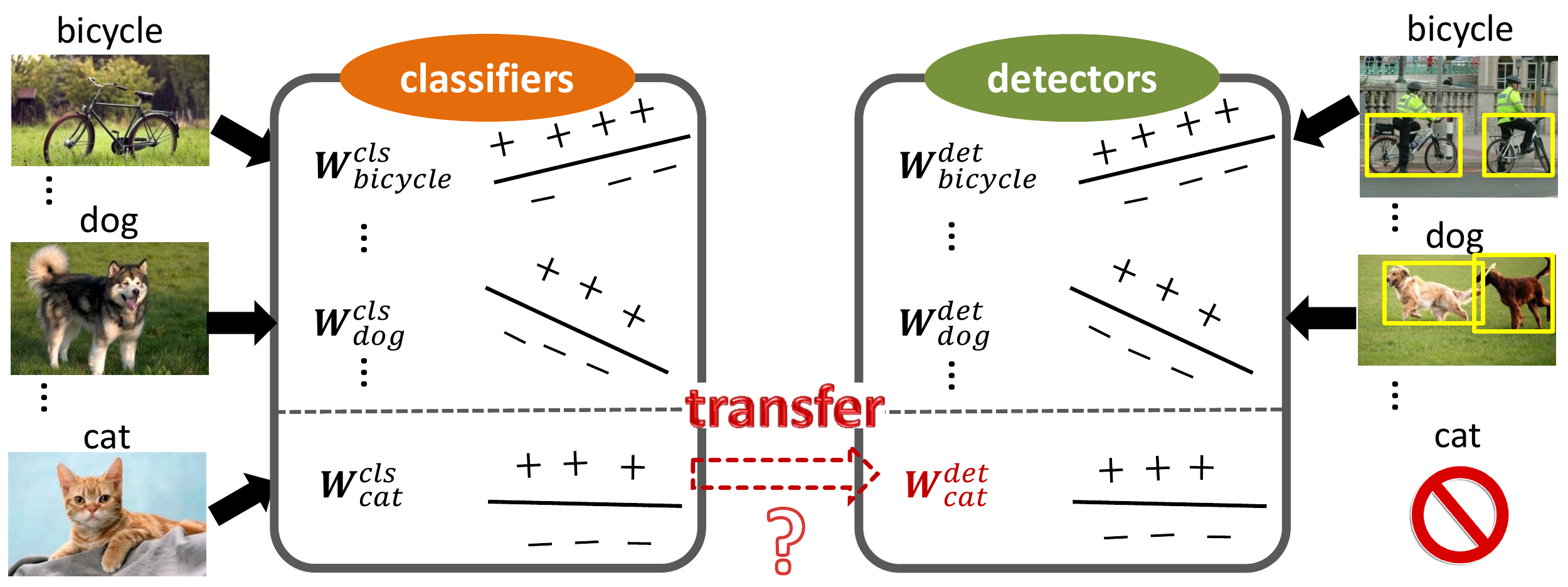}
  \caption{In this work, we consider a dataset containing image-level labels for all the categories, while object-level bounding box annotations are only available for some of the categories (\ie, weakly labeled categories). How can we transform a CNN classification network into a detection network to detect the weakly labeled categories (\eg, the \textit{cat} class)?}
  \label{fig:intro}
\end{figure*}
%%%%%%%%%%%%%%%%Fig%%%%%%%%%%%%%%%%%%%%%%%%%%%%%%%%%%%%%%%%%%

% \subsection{Motivation}
% \label{semi: Motivation}

Although localized object annotations are extremely valuable, the process of manually annotating object bounding boxes is extremely laborious and unreliable, especially for large-scale databases. On the other hand, it is usually much easier to obtain annotations at \emph{image} level (\eg, from user-generated tags on Flickr or Web queries). 
For example, ILSVRC contains image-level annotations for 1,000 categories, while object-level annotations are currently restricted to only 200 categories. One could apply image-level classifiers directly to detect object categories, but this will result in a poor performance as there are differences in the statistical distribution between the training data (whole images) and the test data (localized object instances). 
Previous work by Hoffman~\etal \cite{hoffman2014lsda} addresses this issue, by learning a transformation between CNN classifiers and detectors of object categories with \emph{both} image-level and object-level annotations (``strong'' categories), and applying the transformation to adapt image-level classifiers to object detectors for categories with \emph{only} image-level labels (``weak'' categories). 
Part of this work involves transferring \emph{category-specific} classifier and detector differences of visually similar ``strong'' categories equally to a classifier of a ``weak'' category to form a detector for that category (Fig.~\ref{fig:intro}). We argue that more can potentially be exploited from such similarities in an informed manner to improve detection beyond using the measures solely for nearest neighbor selection (see Section \ref{background}).
Moreover, since there is evidence that deep CNNs trained for image classification also learn proxies to objects and object parts \cite{zhou2014object}, the transformation from CNN classifiers to detectors is reasonable and practicable.

% \subsection{Contribution}
% \label{semi: Contribution}

\begin{figure*}[htbp!]
  \centering
  \includegraphics[width=0.95\linewidth]{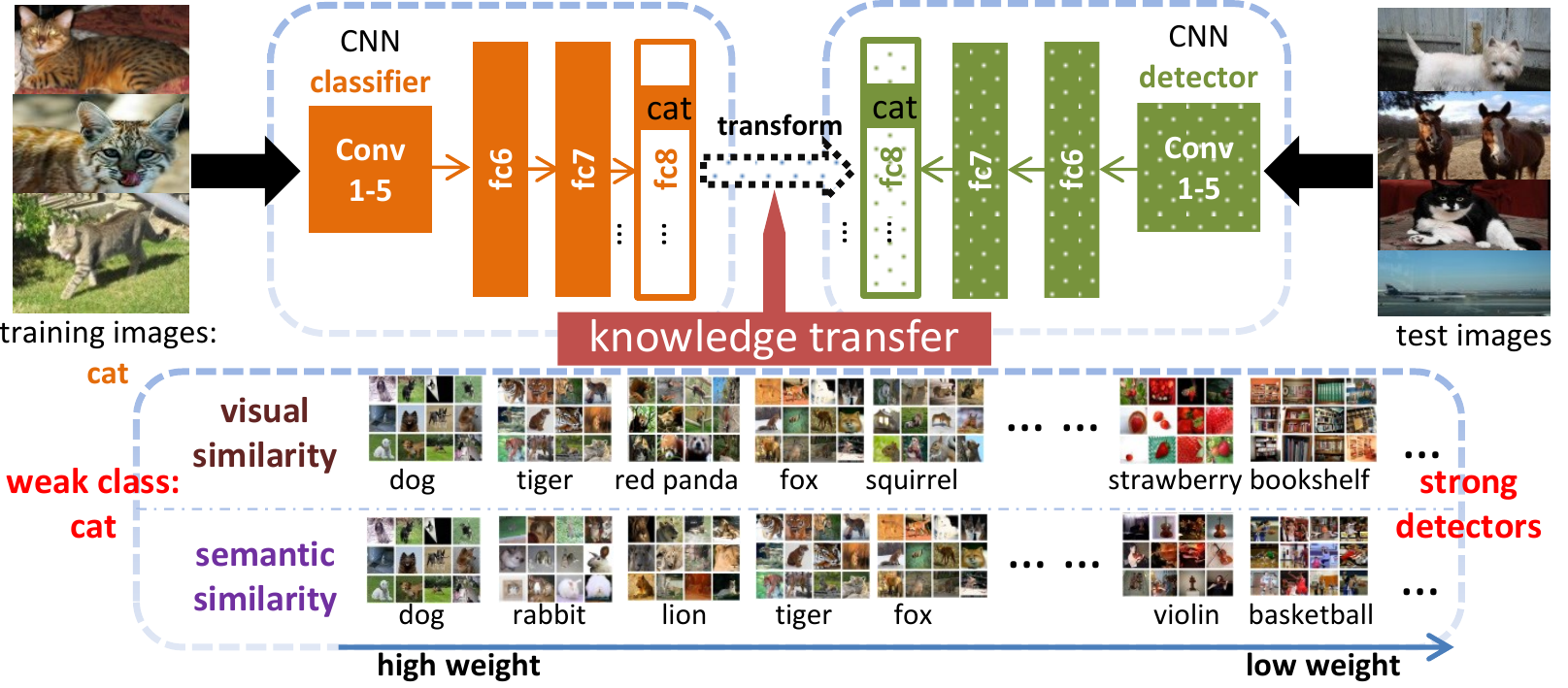}
  \caption{An illustration of our similarity-based knowledge transfer model. The question we investigate %in this paper 
is whether knowledge about object similarities -- visual and semantic -- can be exploited to improve detectors trained in a semi-supervised manner. More specifically, to adapt the image-level classifier (up-left) of a ``weakly labeled'' category (no bounding boxes) into a detector (up-right),  we transfer information about the classifier and detector differences of ``strong'' categories (with image-level and bounding box annotations, bottom of the figure) by favoring categories that are more similar to the target category (\eg, transfer information from \textit{dog} and \textit{tiger} rather than \textit{basketball} or \textit{bookshelf} to produce a \textit{cat} detector).}
  \label{fig:Overview}
\end{figure*}

Our main contribution in this paper is therefore to incorporate external knowledge about object similarities from visual \emph{and} semantic domains in modeling the aforementioned category-specific differences, and subsequently transferring this knowledge for adapting an image classifier to an object detector for a ``weak'' category. Our proposed method is motivated by the following observations: (i) category specific difference exists between a classifier and a detector \cite{girshick2014rcnn, hoffman2014lsda}; (ii) visually and semantically similar categories may exhibit more common transferable properties than visually or semantically dissimilar categories; (iii) visual similarity and semantic relatedness are shown to be correlated, especially when measured against object instances cropped out from images (thus discarding background clutter)~\cite{Deselaers_sim10CVPR}. Intuitively, we would prefer to adapt a \textit{cat} classifier to a \textit{cat} detector by using the category-specific differences between the classifier and the detector of a \textit{dog} rather than of a \textit{violin} or a \textit{strawberry} (Fig.~\ref{fig:Overview}). The main advantage of our proposed method is that knowledge about object similarities can be obtained without requiring further object-level annotations, for example from existing image databases, text corpora and external knowledge bases.

Our work aims to answer the question: can knowledge about visual and semantic similarities of object categories (and the combination of both) help improve the performance of detectors trained in a weakly supervised setting (\ie, by converting an image classifier into an object detector for categories with only image-level annotations)? Our claim is that by exploiting knowledge about objects that are visually and semantically similar, we can better model the category-specific differences between an image classifier and an object detector and hence improve detection performance, without requiring bounding box annotations. We also hypothesize that the combination of both visual and semantic similarities can help further improve the detector performance. Experimental results on the challenging ILSVRC2013 dataset \cite{ILSVRC15} validate these claims, showing the effectiveness of our approach of transferring knowledge about object similarities from both visual and semantic domains to adapt image classifiers into object detectors in a semi-supervised manner.

A preliminary version of this work appeared in \cite{Tang_2016_CVPR}. In this paper, we provide more technical details of our models, introduce a bounding-box post-processing method based on the transferability of regression models, and present extended results with more comparisons.
The rest of the paper is organized as follows. We review related work in Section \ref{Semi-related} and define the semi-supervised object detection problem in Section \ref{definition}.
In Section \ref{approach}, we first review the LSDA framework, and then introduce our two knowledge transferring methods (\ie\ visual similarity based method and semantic similarity based method). We present our experimental results and comparisons in Section \ref{exp}. In Section \ref{semi-concl}, we conclude and describe future direction.

%%%%%%%%%%%%%%%%Fig overview%%%%%%%%%%%%%%%%%%%%%%%%%%%%%%%%%%%%%%%%%%
%%%%%%%%%%%%%%%%%Fig%%%%%%%%%%%%%%%%%%%%%%%%%%%%%%%%%%%%%%%%%%

%------------------------------------------------------------------------
\section{Related Work}
\label{Semi-related}

With the remarkable success of deep CNN on large-scale object recognition
\cite{AlexNet_NIPS2012} in recent years, a substantial number of CNN-based object
detection frameworks~\cite{ Szegedy_NIPS2013, sermanet-iclr-14, girshick2014rcnn, girshick15fastrcnn, SPPnet_TPAMI2015, 
ren15fasterrcnn, Redmon_2016_CVPR, liu2016single} have emerged. However, these object detectors are trained in a fully supervised manner, where bounding box annotations are necessary during training.
The requirement of bounding box annotations hinders the application of these methods in large-scale datasets where training images are weakly annotated. 

\subsection{Weakly Supervised Learning}
\label{related: WSL}

Weakly supervised learning methods for object detection attempt to learn localization cues from image-wide labels indicating the presence or absence of object instances of a category, thus reducing or removing the requirement of bounding box annotations~\cite{Crandall-et-al-ECCV06, Chum07a, Galleguillos_ECCV08, Nguyen-et-al-ICCV09, Siva-et-al-ICCV11, Pandey-et-al-ICCV11, siva2012weakly, Deselaers-et-al-IJCV12, Shi_2013_ICCV, ytangICIP2014}. 
Recently, there have been several studies in CNN-based object detection in a \emph{weakly}-supervised setting~\cite{bilen2014bmvc, songICML14slsvm, Bilen_2015_CVPR, CWang_LCL_TIP2015, Tang_TMM_WSL}, \ie\ using training images with only image-level labels and no bounding boxes. 
The common practice is to jointly learn an appearance model
together with the latent object location from such weak annotations. 
Such approaches only adopt CNN as a feature extractor, and exhaustively mine image regions extracted by region proposal approaches, such as Selective Search \cite{UijlingsIJCV13}, BING \cite{cheng2014bing}, and EdgeBoxes \cite{ZitnickECCV14edgeBoxes}. Oquab \etal \cite{Oquab_2015_CVPR}
develop a weakly supervised CNN end-to-end learning pipeline that learns
from complex cluttered scenes containing multiple objects by explicitly
searching over possible object locations and scales in the image, which
can predict image labels and coarse locations (but not exact bounding boxes) of objects. 
Bilen and Vedaldi \cite{Bilen16} propose a Weakly Supervised Deep Detection Network (WSDNN) method that extends a pre-trained network to a two-stream CNN: recognition and detection. The recognition and detection scores for region proposals are aggregated to predict the object category. 
Zhou \etal \cite{zhou2016cvpr} adopt a classification-trained CNN to learn to localize object by generating Class Activation Maps (CAM) using global average pooling (GAP).
Hoffman \etal \cite{hoffman2014lsda} propose a Large Scale
Detection through Adaptation (LSDA) algorithm that learns the difference
between the CNN parameters of the image classifier and object detector of a
``fully labeled'' category, and transfers this knowledge to CNN classifiers
for categories without bounding box annotated data, turning them into detectors.
For LSDA, auxiliary object-level annotations for a subset of the categories are
required for training ``strong'' detectors. This can be considered a semi-supervised
learning problem (see Section \ref{definition}%for definition
). We improve upon LSDA, by incorporating knowledge about visual and semantic similarities of object categories during the transfer from a classifier to a detector.

\subsection{Transfer Learning}
\label{related: TL}

Another line of related work is to exploit knowledge transfer from various
domains. Transfer learning (TL) \cite{Shao_TL_survey} aims to transfer
knowledge across different domains or tasks. Two general categories of TL
have been proposed in previous work: \emph{homogeneous} TL \cite{Donahue_CVPR2013, Oquab14, hoffman2014lsda} in a single domain but with different data distributions in training and testing sets, and \emph{heterogeneous} TL \cite{Rochan_2015_CVPR, shu_DTNs_MM, zhu2011heterogeneous} across different domains or modalities.
LSDA treats the transfer from classifiers to detectors as a homogeneous TL problem as the data distributions for image classification (whole image features) and object detection (image region features) are different. The adaptation from a classifier to a detector is, however, restricted to %within 
the visual domain. 
Lu \etal \cite{Lu_2017_TNNLS} propose a sparse representation based discriminative knowledge transfer method that leverages relatedness of various source categories with the target category, where only few training samples existed, to enhance learning of the target classifier. 
Rochan and Wang \cite{Rochan_2015_CVPR} propose an appearance  transfer method by transferring semantic knowledge (heterogeneous TL) from familiar 
objects to help localize novel objects in images and videos. 
Singh \etal \cite{krishna-cvpr2016} transfer tracked object boxes from weakly labeled videos to weakly labeled images to automatically generate pseudo ground-truth bounding boxes.
Our work integrates knowledge transfer via both visual similarity (homogeneous TL) and semantic relatedness (heterogeneous TL) to help convert classifiers into detectors.
Frome \etal \cite{Frome2013} present a deep visual-semantic embedding model learned to recognize visual objects using both labeled image data and semantic information collected from unannotated text.
Shu \etal \cite{shu_DTNs_MM} propose a weakly-shared Deep Transfer Network (DTN) that hierarchically learns to transfer semantic knowledge from web texts to images for image classification, building upon Stacked Auto-Encoders \cite{Bengio07_sae}. DTN takes auxiliary text annotations (user tags and comments) and image pairs as input, while our semantic transfer method only requires image-level labels. %as annotation.

\subsection{Semantic Similarity of Text}
\label{related: NLP}

Semantic similarity is a well-explored area within the Natural Language Processing community. The main objective is to measure the distance between the \emph{semantic} meanings of a pair of words, phrases, sentences, or documents. For example, the word ``\textit{car}'' is more similar to ``\textit{bus}'' than it is to ``\textit{cat}''. The two main approaches to measuring semantic similarity are knowledge-based approaches and corpus-based, distributional methods. In the case of knowledge-based approaches, external resources such as thesauri (primarily WordNet~\cite{Fellbaum:98:Wordnet}) or online knowledge bases are used to compute the similarity between the semantic meaning of two terms, for example using path-based similarity~\cite{Leacock:98} or information-content similarity measures~\cite{Resnik:95, Lin:98}. The heavy reliance on knowledge bases, which tend to suffer from issues such as missing words, resulted in the development of \emph{distributional} methods that rely instead on text corpora. In such methods, each term is represented as a \emph{context} vector, and two terms are assumed to have similar vectors if they occur frequently in the same context (e.g. ``\textit{car}'' and ``\textit{truck}'' have similar vectors because they often co-occur with ``\textit{drive}'' and ``\textit{road}''). Such context vectors are more often referred to in recent years as \emph{word embeddings}. Recent advances in word embeddings trained on large-scale text corpora~\cite{Mikolov:13,Pennington:14} have helped progress research in distributional methods to semantic similarity, as it has been observed that semantically related word vectors tend to be close in the embedding space, and that the embeddings capture various linguistic regularities~\cite{Mikolov:13:NAACL} (\textit{King} - \textit{Man} + \textit{Woman} $\approx$ \text{Queen}). As such, we will concentrate on such state-of-the-art word embedding methods to measure the semantic similarity of terms.

%------------------------------------------------------------------------
\section{Task Definition}
\label{definition}
In our semi-supervised learning case, we assume that we have a set of
``fully labeled'' categories and ``weakly labeled'' categories. For the
``fully labeled'' categories, a large number of training images with
both image-level labels and bounding box annotations are available for 
learning the object detectors.
For each of the ``weakly labeled'' categories, we have many training images
containing the target object, but we do not have access to the exact locations
of the objects. This is different from the semi-supervised learning proposed
in  previous work \cite{MisraExemplarSelection, Rosenberg_WACV05, Yang_cvpr13},
where typically a small amount of fully labeled data with a large amount of
weakly labeled (or unlabeled) data are provided for each category. In our
semi-supervised object detection scenario, the objective is to transfer the
trained image classifiers into object detectors on the ``weakly labeled''
categories.

%------------------------------------------------------------------------
%\section{Our Approach}
\section{Similarity-based Knowledge Transfer}
\label{approach}

We first describe the Large Scale Detection through Adaptation (LSDA) framework~\cite{hoffman2014lsda}, upon which our proposed approach is based (Section  \ref{background}). We then describe our proposed knowledge transfer models with the aim of improving LSDA. Two knowledge domains are explored: (i) visual similarity (Section \ref{visual_sim}); (ii) semantic relatedness (Section \ref{semantic_sim}). Next, we combine both models to obtain our mixture transfer model, as presented in Section \ref{fusion_sim}. Finally, we propose to transfer the knowledge to bounding-box regression from fully labeled categories to weakly labeled categories in Section \ref{bb_reg}.

\subsection{Background on LSDA}
\label{background}

% Let $\mathcal{D}=\{C_1, C_2, \ldots, C_m, \ldots, C_K\}$ be the $K$
Let $\mathcal{D}$ be the dataset of $K$
categories to be detected. One has access to both image-level and bounding box annotations
only for a set of $m$ ($m\ll K$) ``fully labeled'' categories, denoted
% as: $\mathcal{B}=\{C_1, C_2, \ldots, C_m\}$, and also has image level
as $\mathcal{B}$, but only image-level
annotations for the rest of the categories, namely ``weakly labeled''
% categories, denoted as: $\mathcal{A}=\{C_{m+1}, C_{m+2}, \ldots, C_K\}$.
categories, denoted as $\mathcal{A}$.
Hence, a set of $K$ image classifiers can be trained on the whole dataset
$\mathcal{D}$ ($\mathcal{D}=\mathcal{A}\cup \mathcal{B}$), but only $m$
object detectors (from $\mathcal{B}$) can be learned according to the
availability of bounding box annotations. The LSDA algorithm learns to
convert $(K-m)$ image classifiers (from $\mathcal{A}$) into their corresponding
object detectors through the following steps:

\noindent \textbf{Pre-training:} First, an 8-layer (5 convolutional
layers and 3 fully-connected ($fc$) layers) \emph{Alex-Net} \cite{AlexNet_NIPS2012}
CNN is pre-trained on the ImageNet Large Scale Visual Recognition Challenge
(ILSVRC) 2012 classification dataset \cite{ILSVRC15}, which contains 1.2
million images of 1,000 categories.

\noindent \textbf{Fine-tuning for classification:} The final weight
layer (1,000 linear classifiers) of the pre-trained CNN is then replaced with
$K$ linear classifiers. This weight layer is randomly initialized and the
whole CNN is then fine-tuned on the dataset $\mathcal{D}$. This produces a
classification network that can classify $K$ categories ({\it i.e.}, $K$-way
softmax classifier), given an image or an image region as input.

\noindent \textbf{Category-invariant adaptation:} Next, the classification network is fine-tuned into a detector with bounding boxes of $\mathcal{B}$ as input, using the R-CNN \cite{girshick2014rcnn} framework. 
%Next, layers 1-7 and the
%output layer $fc8_\mathcal{B}$ of the whole classification network are
%fine-tuned on $\mathcal{B}$ using R-CNN \cite{girshick2014rcnn} to 
%transfer deep features from a classifier to a detector by fine-tuning. Here, labeled region proposals, such as
%Selective Search \cite{UijlingsIJCV13}, are used. 
% We refer the reader to \cite{girshick2014rcnn} for more details concerning R-CNN framework. 
As in R-CNN, a background class ($fc8_\mathcal{BG}$) is added
to the output layer and fine-tuned using bounding boxes from a region proposal algorithm, \eg, Selective Search \cite{UijlingsIJCV13}. The $fc8$ layer parameters
are category \emph{specific}, with 4,097 weights ($fc7$ output: 4,096, plus a
bias term) in each category, while the parameters of layers 1-7
are category \emph{invariant}. Note that object detectors are not able
to be directly trained on $\mathcal{A}$%using R-CNN
, since
the fine-tuning and training process requires bounding box annotations.
Therefore, at this point, the category specific output layer $fc8_\mathcal{A}$
stays unchanged. The variation matrix of $fc8_\mathcal{B}$ after
fine-tuning is denoted as $\Delta_\mathcal{B}$.

\noindent \textbf{Category-specific adaptation:} Finally, each classifier
of categories $j \in \mathcal{A}$ is adapted into a corresponding detector
by learning a category-specific transformation of the model parameters.
This is based on the assumption that the difference between classification
and detection of a target object category has a positive correlation with
those of similar (close) categories. The transformation is computed by
adding a bias vector to the weights of $fc8_\mathcal{A}$. This bias vector
for category $j$ is measured by the average weight change of its $k$
nearest neighbor categories in set $\mathcal{B}$, from classification
to detection.
\begin{equation}
\label{weights_func}
\forall j\in \mathcal{A}: \overrightarrow{w_j^d} = \overrightarrow{w_j^c} + \frac{1}{k}\sum_{i=1}^k\Delta_{\mathcal{B}_i^j}
\end{equation}
where $\Delta_{\mathcal{B}_i^j}$ is the $fc8$ weight variation of the
$i^{th}$ nearest neighbor category in set $\mathcal{B}$ for category
$j \in \mathcal{A}$. $\overrightarrow{w^c}$ and $\overrightarrow{w^d}$
are, respectively, $fc8$ layer weights for the fine-tuned classification
and the adapted detection network. The nearest neighbor categories are
defined as those with nearest $L_2$-\emph{norm} (Euclidean distance)
of $fc8$ weights in set $\mathcal{B}$.

The fully adapted network is able to detect all $K$ categories in
test images. In contrast to R-CNN, which trains SVM classifiers on the
output of the $fc7$ layer %({\it i.e.}, 4,096-dimensional feature vectors), and 
followed by bounding box regression on the extracted features from
the $pool5$ layer of all region proposals, LSDA directly outputs the score
of the softmax ``detector'', and subtracts the background score from this
as the final score. This results in a small drop in performance, but
enables direct adaptation from a classification network into a
detection network on the ``weakly labeled'' categories, and significantly reduces
the training time. 
%from around 3 days to roughly 5.5 hours on the ILSVRC2013 detection dataset.

Hoffman \etal \cite{hoffman2014lsda} demonstrated that
the adapted model yielded a 50\% relative mAP (mean average precision)
boost for detection over the classification-only framework on the
``weakly labeled'' categories of the ILSVRC2013 detection dataset (from 10.31\% to
16.15\%). %They also showed that the category-specific adaptation procedure
%(final step in LSDA) barely contributed to the performance improvement
%(16.15\% with {\it vs.} 15.85\% without this step), indicating that the most important improvement comes from adapting the feature 
%representation (layers 1-7) as well as introducing a background class,
%while the least important is adapting the category specific parameters
%($fc8$ weights). 
They also showed that category-specific adaptation (final LSDA step) contributes least to the performance improvement (16.15\% with {\it vs.} 15.85\% without this step), with the other features (adapted layers 1-7 and background class) being more important. However, we found that by properly adapting this layer, a significant
boost in performance can be achieved: an mAP of
22.03\% can be obtained by replacing
the semi-supervised $fc8_\mathcal{A}$ weights with their corresponding supervised
network weights and leaving the other parameters fixed. Thus, we believe that adapting this layer in an informed manner, such as making better use of knowledge about object similarities, will help improve detection. 

In the next subsections, we will introduce
our knowledge transfer methods using two different kinds of similarity
measurements to select the nearest categories and weight them accordingly to better adapt the $fc8$
layer, which can efficiently convert an image classifier into an object
detector for a ``weakly labeled'' category.

%-------------------------------------------------------------------------
\subsection{Knowledge Transfer via Visual Similarity}
\label{visual_sim}

%in order to transfer an image
%classifier into an object detector based on the similar %learn the difference
Intuitively, the object detector of an object category may be more similar
to those of visually similar categories than of visually distinct categories. For
example, a cat detector may approximate a dog detector better than
a strawberry detector, since cat and dog are both mammals sharing common attributes in terms of shape (both have four legs, two ears, two eyes, one tail) and texture
(both have fur). Therefore, given a ``fully labeled'' dataset $\mathcal{B}$
and a ``weakly labeled'' dataset $\mathcal{A}$, our objective is
to model the visual similarity between each category
$j\in \mathcal{A}$ and all the other categories in $\mathcal{B}$, and to transfer this knowledge for transforming classifiers into detectors for $\mathcal{A}$.

\noindent \textbf{Visual similarity measure:} Visual similarity measurements are often obtained by computing the distance between feature distributions such as %discriminative feature representations for each image by taking the output of 
the $fc6$ or $fc7$ output of a CNN, or in the case of LSDA the $fc8$ layer parameters. %(4,096-dimensional vector).
In our work, we instead forward propagate an image through the whole
fine-tuned classification network (created by the second step in Section
\ref{background}) to obtain a $K$-dimensional classification score vector. 
This score vector encodes the probabilities of an image being each of the $K$ object
categories. Consequently, for all the positive images of an object
category $j\in \mathcal{A}$, we can directly accumulate the scores of
each dimension, on a balanced validation dataset. We assume that the normalized
accumulated scores (range [0,1]) imply the similarities between category
$j$ and other categories: the larger the score, the more it  visually
resembles category $j$. This assumption is supported by the analysis
of deep CNNs \cite{agrawal14analyzing, jia2014caffe, ZeilerECCV2014}:
CNNs are apt to confuse visually similar categories, on which they
might have higher prediction scores. The visual similarity (denoted 
$s_v$) between a ``weakly labeled'' category $j\in \mathcal{A}$ and
a ``fully labeled'' category $i\in \mathcal{B}$ is defined as:
\begin{equation}
\label{vd_func}
    s_v(j,i) \propto \frac{1}{N}{\sum_{n=1}^{N}\text{$CNN_{softmax}$}(I_n)_i}
\end{equation}
where $I_n$ is a positive image from category $j$ of the validation set
of $\mathcal{A}$, $N$ is the number of positive images for this category,
and $\text{\emph{$CNN_{softmax}$}}(I_n)_i$ is the $i^{th}$  CNN output of the softmax
layer on $I_n$, namely, the probability of $I_n$ being category $i \in \mathcal{B}$ as
predicted by the fine-tuned classification network. $s_v(j,i)\in [0,1]$
is the degree of similarity after normalization on all the categories
in $\mathcal{B}$.

Note that we adopt the $fc8$ outputs (classification scores) since most of the computation is integrated into the end-to-end \emph{Alex-Net} framework except for the accumulation of classification scores in the end, saving the extra effort otherwise required for distance computation if $fc6$ or $fc7$ outputs were to be used.
The idea of using the $L_2$ distance of the $fc8$ weights (linear classifier parameters) as a visual similarity measurement in LSDA is closely related to ours. However, in addition to the $fc8$ weights, our visual similarity measurement is assumed to leverage the powerful and supplementary feature representations generated by the prior layers of the neural network by combining both, given the fact that the $fc8$ outputs are obtained by taking the inner-product of $fc7$ outputs (visual features) and $fc8$ weights. Experimental results in Section \ref{exp} validate this intuition. 

\noindent \textbf{Weighted nearest neighbor scheme: } Using Eq.\ \eqref{weights_func}, we can transfer the model parameters based
on a category's %average weight variations of its 
$k$ nearest neighbor categories selected by
Eq.\ \eqref{vd_func}. This allows us to directly compare our visual similarity measure to that of LSDA which uses the Euclidean distance between the $fc8$ parameters. An alternative to Eq.\ \eqref{weights_func} is to consider a \emph{weighted} nearest neighbor scheme, where weights can be assigned to different categories based on how visually similar they are to the target object category. This is intuitive, as different categories will have varied degrees of similarity to a particular class, and some categories may have only a few (or many) visually similar classes. Thus, we modify Eq.\ \eqref{weights_func} and define the transformation via visual similarity based on the proposed weighted nearest neighbor scheme as:  

%This simple approach can be directly compared with the
%nearest category selection scheme in LSDA, to justify that our visual
%similarity measure is superior to consider the Euclidean distance between
%the $fc8$ parameters. 
%Except for that, we can arrange different weights
%to different categories according to their visual similarities with the
%target object class, due to the fact that different categories may have
%distinct relationships (different degrees of similarity) with the target
%object class, and some categories may have few (or plenty) visually similar
%classes. Hence, our transformation form via visual similarity is described
%as follows:

\begin{equation}
\label{weights_func_ours}
\forall j\in \mathcal{A}: {\overrightarrow{w_j^d}}_v = \overrightarrow{w_j^c} + \sum_{i=1}^m s_v(j,i)\Delta_{\mathcal{B}_i^j}
\end{equation}
It is worth noting that %the average $k$ nearest neighbors (Eq.\ \eqref{weights_func})
Eq.\ \eqref{weights_func} is a special case of Eq.\ \eqref{weights_func_ours}, where $m=k$ and $s_v(j,i)=1/k$.

%%%%%%%%%%%%%%%%Fig relationship%%%%%%%%%%%%%%%%%%%%%%%%%%%%%%%%%%%%%%%%%%
\begin{figure*}[htbp!] % Fig relationship
  \centering
  \includegraphics[width=\linewidth]{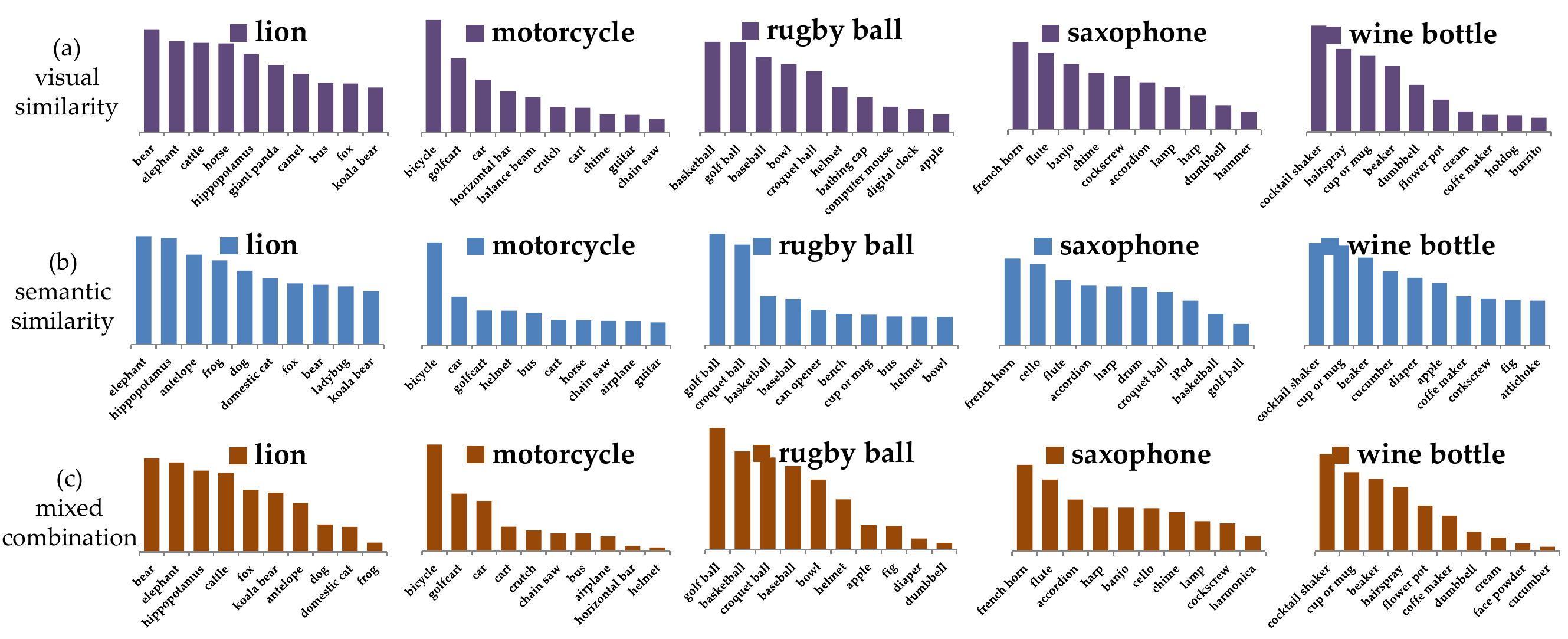}
  % \vspace{-2mm}
  \caption{Some example visualizations of (a) visual similarity (first row in the figure), (b) semantic similarity (middle row) and (c) mixture similarity (last row) between a target ``weakly labeled'' category and its source categories from which knowledge is to be transferred. For each target category, the top-10 weighted nearest neighbor categories are shown. The magnitude of each column bar shows the relative weight (degree of similarity $s_v$, $s_s$, $s$ in Eq. \eqref{eq_fusion}, where $\alpha$ is set to 0.6).}
  \label{fig:relation}
  % \vspace{-3mm}
\end{figure*}
%%%%%%%%%%%%%%%%Fig%%%%%%%%%%%%%%%%%%%%%%%%%%%%%%%%%%%%%%%%%%

%-------------------------------------------------------------------------

\subsection{Knowledge Transfer via Semantic Relatedness}
\label{semantic_sim}
%The main shortage of our knowledge transfer via visual similarity lies on
%the contextual information! maybe the context is very similar.
Following prior work \cite{Deselaers_sim10CVPR, Rochan_2015_CVPR, rohrbach10cvpr},
we observe that visual similarity is correlated with semantic relatedness.
According to \cite{Deselaers_sim10CVPR}, this relationship is particularly
strong when measurements are focused on the category instances themselves,
ignoring image backgrounds. This observation is quite intriguing for object
detection, where the main focus is on the target objects themselves.
Hence, we draw on this fact and propose transferring knowledge
from the natural language domain to help improve semi-supervised object
detection.

\noindent \textbf{Semantic similarity measure:} %Semantic similarity is a well-explored area within the Natural Language Processing community. The two main approaches to measuring semantic similarity are knowledge-based approaches and corpus-based, distributional methods. 
%Recent advances in word embeddings trained on large-scale text corpora~\cite{Mikolov:13,Pennington:14} have helped progress research in this area, as it has been observed that semantically related word vectors tend to be close in the embedding space, and that the embeddings capture various linguistic regularities~\cite{Mikolov:13:NAACL}. Thus,% 
As mentioned in Section~\ref{related: NLP}, we use word embeddings to represent each category and to measure the semantic similarity between categories. Each of the $K$ categories is embedded as a word vector, more specifically a 300-dimensional word2vec embedding~\cite{Mikolov:13}. 
Since each category is a WordNet~\cite{Fellbaum:98:Wordnet} synset, we represent each category as the sum of the word vectors for each term in its synset, normalized to unit vector by its $L_2$-\emph{norm}.
Out-of-vocabulary words are addressed by attempting to match case variants of the words (lowercase, Capitalized), \eg, ``aeroplane'' is not in the vocabulary, but ``Aeroplane'' is. Failing that, we represent multiword phrases by the sum of the word vectors of each in-vocabulary word of the phrase, normalized to unit vector (``baby''+``bed'' for \emph{baby bed}). In several cases, we also augment synset terms with any category label defined in ILSVRC2013 that is not among the synset terms defined in WordNet (e.g. ``bookshelf'' for the WordNet synset \textit{bookcase}, and  ``tv'' and ``monitor'' for \textit{display}). 
%\yuxing{the category name is ``tv \textbf{or} monitor''?} \josiah{Yes, I am aware of this one. In WordNet, this synset only has ``display'' and ``video display'' as their words, so I added ``tv'' AND ``monitor'' (as two sthe separate words) to the vocabulary, if that makes sense? We can remove the second example if it's confusing} \yuxing{OK. I understand now. I think we can keep it.}
% * <limingchen69@gmail.com> 2015-11-04T11:06:42.577Z:
%
% > Out-of-vocabulary words are addressed by attempting to match case variants of the words (lowercase, Capitalized). Failing that, we represent multiword phrases by the sum of the word vectors of each in-vocabulary word of the phrase, normalized to unit vector.
%
% do we have these cases in this work ? have we cases of multiword phrases ?
%
% ^.

Word embeddings often conflate multiple senses of a word into a single vector, leading to an issue with polysemous words. We observed this with many categories, for example \textit{seal} (animal) is close to \textit{nail} and \textit{tie} (which, to further complicate matters, is actually meant to refer to its clothing sense); or the stationery \textit{ruler} being related to \textit{lion}. Since ILSVRC2013 categories are actually WordNet synsets, it makes perfect sense to exploit WordNet to help disambiguate the word senses. Thus, we integrate corpus-based representations with semantic knowledge from WordNet, by using AutoExtend~\cite{Rothe:15} to encode the categories as \emph{synset embeddings} in the original word2vec embedding space. AutoExtend exploits the interrelations between synsets, words and lexemes to learn an auto-encoder based on these constraints, as well as constraints on WordNet relations such as hypernyms (encouraging \textit{poodle} and \textit{dog} to have similar embeddings). We observed that AutoExtend has indeed helped form better semantic relations between the desired categories: \textit{seal} is now clustered with other animal categories like \textit{whale} and \textit{turtle}, and the nearest neighbors for \textit{ruler} are now \textit{rubber eraser}, \textit{power drill} and \textit{pencil box}. In our detection experiments (Section~\ref{exp}), we found that while the `naive' word embeddings performed better than the baselines, the synset embeddings yielded even better results. Thus, we
concentrate on reporting the results of the latter.%only report the results for the latter.

%After each category $j\in \mathcal{A}$ and $i\in \mathcal{B}$ has been represented by a feature vector, we can calculate the $L_2$-\emph{norm} of each pair $d_s(j,i)$ as their semantic distance.
%The semantic distance of each pair, $d_s(j,i)$ $: \forall j\in \mathcal{A} \wedge \forall i\in \mathcal{B}$, is computed as the $L_2$-\emph{norm} of the synset embeddings.

We represent each category $j\in \mathcal{A}$ and $i\in \mathcal{B}$ with their synset embeddings, and compute the $L_2$-\emph{norm} of each pair $d_s(j,i)$ as their semantic distance. The semantic similarity $s_s(j,i)$ is inversely proportional to $d_s(j,i)$. We then transfer the semantic knowledge to the appearance model using Eq.\ \eqref
{weights_func_ours} or its special case Eq.\ \eqref{weights_func} as before. %of $k$ nearest neighbor categories. 

%On the other hand, according to \cite{Rochan_2015_CVPR}, 
As our semantic representations are in the form of vectors, we explore an alternative similarity measure as used in \cite{Rochan_2015_CVPR}. We assume
that each vector of a ``weakly labeled'' category $j\in \mathcal{A}$
(denoted as $v_j$) can be approximately represented by a linear
combination of all the $m$ word vectors in $\mathcal{B}$:
$v_j \approx \Gamma_j V$, where $V = [v_1; v_2; \ldots; v_i; \ldots; v_m]$,
and $\Gamma_j = [\gamma_j^1, \gamma_j^2, \ldots, \gamma_j^i, \ldots, \gamma_j^m]$
is a set of coefficients of the linear combination. We are motivated
to find the solution $\Gamma_j^\star$ which contains as few non-zero
components as possible, since we tend to reconstruct category $j$
with fewer categories from $\mathcal{B}$ (sparse representation).
This optimal solution $\Gamma_j^\star$ can be formulated as the
following optimization:
\begin{equation}
\label{recon_func}
\Gamma_j^\star = \argmin_{\Gamma_j > 0} (\|v_j - \Gamma_j V\|_2 + \lambda \|\Gamma_j\|_0)
\end{equation}
Note that $\Gamma_j > 0$ is a positive constraint on the coefficients,
since negative components of sparse solutions for semantic transferring
are meaningless: we only care about the most similar categories and not
dissimilar categories. We solve Eq.\ \eqref{recon_func} by using the
positive constraint matching pursuit (PCMP) algorithm \cite{gao_pcmp_cbmi12}.
Therefore, the final transformation via semantic transferring is formulated as:
\begin{equation}
\label{seman_func_ours}
\forall j\in \mathcal{A}: {\overrightarrow{w_j^d}}_s = \overrightarrow{w_j^c} + \sum_{i=1}^m s_s(j,i)\Delta_{\mathcal{B}_i^j}
\end{equation}
where $s_s(j,i) = \gamma_j^i$ in the sparse representation case.

%-------------------------------------------------------------------------
\subsection{Mixture Transfer Model}
\label{fusion_sim}
We have proposed two different knowledge transfer models. Each of them can be
integrated into the LSDA framework independently. In addition, since we
consider the visual similarity at the whole image level and the semantic
relatedness at object level, they can be combined simultaneously to provide
complementary information. We use a simple but very effective combination of
the two knowledge transfer models as our final mixture transfer model. Our
mixture model is a linear combination of the visual similarity and the semantic
similarity:
\begin{equation}
\label{eq_fusion}
s = \intersec [\alpha s_v + (1-\alpha)s_s]
\end{equation}
where $\intersec[\cdot]$ is a function that takes the intersection of
cooccurring categories between visual and sparse semantic related
categories. $\alpha \in [0,1]$ is a parameter used to control the relative
influence of the two similarity measurements. $\alpha$ is set to 1
when only considering visual similarity transfer, and 0 for the
semantic similarity transfer. We will analyze this parameter in %the
%experimental evaluation (See 
Section \ref{results_eval}.%).

\subsection{Transfer on Bounding-box Regression}
\label{bb_reg}
The detection windows generated by the region based detection models are the highest scoring proposals (\eg, Selective Search). In order to improve localization performance, a bounding-box regression stage \cite{ girshick2014rcnn} is commonly adopted to post-process the detection windows. This process needs bounding box annotations in training the regressors, which is an obstacle for ``weakly labeled'' categories in our case. Hence, we propose to transfer the class-specific regressors from ``fully labeled'' categories to ``weakly labeled'' categories based on the aforementioned similarity measures.

To train a regressor for each ``fully labeled'' category, we select a set of N training pairs $\{(\vec{P^i}, \vec{G^i})\}_{i=1,\ldots, N}$, where $\vec{P^i} = (P_x^i, P_y^i, P_w^i, P_h^i)$ is a vector indicating the center coordinates ($P_x^i, P_y^i$) of proposal $P^i$ together with $P^i$'s width and height ($P_w^i, P_h^i$).
$\vec{G^i} = (G_x^i, G_y^i, G_w^i, G_h^i)$ is the corresponding ground-truth bounding box. Except where needed to avoid confusion we omit the super-script $i$.
The goal is to learn a mapping function $f(P) = (f_x(P), f_y(P), f_w(P), f_h(P))$ which maps a region proposal $P$ to a ground-truth window $G$. Each function within $f(P)$ is modeled as a linear function of the $pool5$ features (namely the feature map after the last convolutional and pooling block of the ConvNet): $f(P) = \mathbf{w}^T_*F_5(P)$, where $\mathbf{w}_*$ is a vector of learnable parameters, $F_5(P)$ is the $pool5$ feature of region proposal $P$. $\mathbf{w}_*$ can be learned by optimizing the following least squares objective function:
\begin{equation}
\label{least_sq_func}
\mathbf{w}_* = \argmin_{\hat{\mathbf{w}}_*} \sum_{i=1}^N(\hat{\mathbf{w}}^T_*F_5(P^i) - t_*^i)^2 + \lambda_0\|\hat{\mathbf{w}}_*\|^2
\end{equation}
where $t_* = (t_x, t_y, t_w, t_h)$ is the regression target for the training pair ($P, G$) defined as: 
\begin{equation}
\label{reg_target}
  \begin{array}{r c l}
  t_x & = & (G_x - P_x)/P_w, \\
  t_y & = & (G_y - P_y)/P_h, \\
  t_w & = & \log(G_w/P_w), \\
  t_h & = & \log(G_h/P_h). \\
  \end{array}
\end{equation}
The first two equations specify a scale-invariant translation of the center of the bounding box, while the remaining two specify the log-space translation of the width and height of the bounding box. After learning the parameters of the transformation function, a detection window (region proposal) $P$ can be transformed into a new prediction $\hat{P} = (\hat{P}_x, \hat{P}_y, \hat{P}_w, \hat{P}_h)$ by applying:
\begin{equation}
\label{reg_trans}
  \begin{array}{r c l}
  \hat{P}_x & = & P_x + P_w f_x(P), \\
  \hat{P}_y & = & P_y + P_h f_h(P), \\
  \hat{P}_w & = & P_w \exp (f_w(P)), \\
  \hat{P}_h & = & P_h \exp (f_h(P)). \\
  \end{array}
\end{equation}

The training pair ($P, G$) is selected if the proposal $P$ has maximum IoU overlap with ground-truth bounding box $G$. The pair ($P, G$) is discarded if the maximum IoU overlap is less than a threshold (which is set to be 0.6 using a validation set).

For a ``weakly labeled'' category $j$, the transformation function cannot be explicitly learned due to the absence of ground-truth bounding boxes. However, we can still transfer this knowledge from similar categories in the ``fully labeled'' subset $\mathcal{B}$: 

\begin{equation}
\label{eq_reg_trans}
\forall j\in \mathcal{A}: \mathbf{w}_j = \sum_{i=1}^m s_*\mathbf{w}_i
\end{equation}
where $s_*$ indicates any one of the aforementioned similarity measures.
%------------------------------------------------------------------------
\section{Experiments}
\label{exp}

%%%%%%%%%%%%%%%%%%%%%%%%%%%%%%%%table%%%%%%%%%%%%%%%%%%%%%%%%%%%%%%%%%%%%%%%%%%%%%%
\begin{table*}[htbp] \fontsize{9pt}{9pt}\selectfont
\renewcommand{\arraystretch}{1.4}
\caption{Detection mean average precision (mAP) on ILSVRC2013 val2. The first row shows the basic performance of directly using all classification parameters for detection, without adaptation or knowledge transfer ({\it i.e.}, weakly supervised learning). The last row shows results of an oracle detection network which assumes that bounding boxes for all 200 categories are available ({\it i.e.}, supervised learning). The second row shows the baseline LSDA results using only feature adaptation. Rows 3-5 show the performance of LSDA for adapting both the feature layers (layer 1-7) and the class-specific layer (layer 8), by considering different numbers of neighbor categories. Rows 6-8, 9-12 and row 13 show the results of our visual transfer, semantic transfer and mixture transfer model, respectively. Row 14 shows our results after bounding-box regression. For all methods, the same ``\textit{\textbf{AlexNet}}'' CNN is adopted.} %Improvement over the corresponding baseline method is shown in red.} 
\label{table-results}
\begin{center}
\begin{tabular}{l c||c c|c}
  \hline
  Method & \tabincell{c}{Number of\\ Nearest Neighbors} & \tabincell{c}{mAP on $\mathcal{B}$: \\ ``Fully labeled''\\ 100 Categories } & \tabincell{c}{mAP on $\mathcal{A}$:\\ ``Weakly labeled''\\ 100 Categories}  & \tabincell{c}{mAP on $\mathcal{D}$:\\All\\ 200 Categories}\\ \hline
  %%%%%%%%%%%%%%%%%%%%%%%%%%%%%%%%%%%%%%%%%%%%%%%%%%%%%%%%%%%%%%%%%%%%%%%%%%%%%%%%%%%%%%%%%%%%%%
  Classification Network                    &-              &12.63 &10.31 &11.90\\ \hline
  LSDA (only class invariant adaptation)    &-              &27.81 &15.85 &21.83\\ \hline
  %%%%%%%%%%%%%%%%%%%%%%%%%%%%%%%%%%%%%%%%%%%%%%%%%%%%%%%%%%%%%%%%%%%%%%%%%%%%%%%%%%%%%%%%%%%%%%
%                                             &Average - 5    &28.12 &15.97 &22.05\\
%   LSDA (class invariant \& specific adapt.) &Average - 10   &27.95 &16.15 &22.05\\
%                                             &Average - 100  &27.91 &15.96 &21.94\\ \hline \hline
%   %%%%%%%%%%%%%%%%%%%%%%%%%%%%%%%%%%%%%%%%%%%%%%%%%%%%%%%%%%%%%%%%%%%%%%%%%%%%%%%%%%%%%%%%%%%%%%
%                                             &Average - 5    &27.99 &{17.42 \textcolor[rgb]{1.00,0.00,0.00}{$\uparrow$1.45}} & 22.71\\
%   Ours (visual transfer)                    &Average - 10   &27.89 &{17.62 \textcolor[rgb]{1.00,0.00,0.00}{$\uparrow$1.47}} & 22.76\\
%                                             &Weighted - 100 &28.30 &{19.02 \textcolor[rgb]{1.00,0.00,0.00}{$\uparrow$3.17}} & 23.66\\
%   \hline
%   %%%%%%%%%%%%%%%%%%%%%%%%%%%%%%%%%%%%%%%%%%%%%%%%%%%%%%%%%%%%%%%%%%%%%%%%%%%%%%%%%%%%%%%%%%%%%%
%                                             &Average - 5    &28.01 &{17.32 \textcolor[rgb]{1.00,0.00,0.00}{$\uparrow$1.35}} & 22.67\\
%   Ours (semantic transfer)                  &Average - 10   &28.00 &{16.67 \textcolor[rgb]{1.00,0.00,0.00}{$\uparrow$0.52}} & 22.31\\
%                                             &Weighted - 100 &28.14 &{18.32 \textcolor[rgb]{1.00,0.00,0.00}{$\uparrow$2.47}} & 23.28\\
%                                     &Sparse rep. - $\leq$20 &28.18 &{19.04 \textcolor[rgb]{1.00,0.00,0.00}{$\uparrow$2.89}} & 23.66\\
%   \hline
%  

                                          &avg/weighted - 5 &28.12 / -- &15.97 / 16.12 &22.05 / 22.12\\
LSDA (class invariant \& specific adapt) &avg/weighted - 10 &27.95 / -- &16.15 / 16.28 &22.05 / 22.12\\
                                          &avg/weighted - 100 &27.91 / -- &15.96 / 16.33 &21.94 / 22.12
                                          \\ \hline \hline
  %%%%%%%%%%%%%%%%%%%%%%%%%%%%%%%%%%%%%%%%%%%%%%%%%%%%%%%%%%%%%%%%%%%%%%%%%%%%%%%%%%%%%%%%%%%%%%
                                            &avg/weighted - 5    &27.99 / -- &17.42 / 17.59 & 22.71 / 22.79\\
  \textbf{Ours (visual transfer)}                    &avg/weighted - 10   &27.89 / -- &17.62 / 18.41  & 22.76 / 23.15\\
                                            &avg/weighted - 100 &28.30 / -- &17.38 / 19.02  & 22.84 / 23.66\\
  \hline
  %%%%%%%%%%%%%%%%%%%%%%%%%%%%%%%%%%%%%%%%%%%%%%%%%%%%%%%%%%%%%%%%%%%%%%%%%%%%%%%%%%%%%%%%%%%%%%
                                            &avg/weighted - 5    &28.01 / -- &17.32 / 17.53 & 22.67 / 22.77\\
  \textbf{Ours (semantic transfer)}                  &avg/weighted - 10   &28.00 / -- &16.67 / 17.50 & 22.31 / 22.75\\
                                            &avg/weighted - 100 &28.14 / -- &17.04 / 18.32  & 23.23 / 23.28\\
                                    &Sparse rep. - $\leq$20 &28.18 &19.04 & 23.66\\
  \hline
 
%%%%%%%%%%%%%%%%%%%%%%%%%%%%%%%%%%%%%%%%%%%%%%%%%%%%%%%%%%%%%%%%%%%%%%%%%%%%%%%%%%%%%%%%%%%%%%
  \textbf{Ours (mixture transfer ) }            &-       &28.04 &{\textbf{20.03 \textcolor[rgb]{1.00,0.00,0.00}{$\uparrow$3.88}}} & 24.04\\
    \textbf{Ours (mixture transfer + BB reg.)}  &-    &31.85 &\textbf{21.88} & 26.87\\
  \hline \hline
  %%%%%%%%%%%%%%%%%%%%%%%%%%%%%%%%%%%%%%%%%%%%%%%%%%%%%%%%%%%%%%%%%%%%%%%%%%%%%%%%%%%%%%%%%%%%%%
  Oracle: Full Detection Network (no BB reg.)           &-              &29.72 &26.25 &28.00\\
  Oracle: Full Detection Network (BB reg.)           &-              &32.17 &29.46 &30.82\\ \hline
\end{tabular}
\end{center}
% space{-3mm}
\end{table*}
%%%%%%%%%%%%%%%%%%%%%%%%%%%%%%%%%%%%%%%%%%%%%%%%%%%%%%%%%%%%%%%%%%%%%%%%%%%%%%%%%%%%%%

%-------------------------------------------------------------------------
\subsection{Dataset Overview}
\label{dataset}

We investigate the proposed knowledge transfer models for large scale semi-supervised object detection on the ILSVRC2013 detection dataset
covering 200 object categories. %This dataset covers
%200 object categories, and it is split into three subsets: training,
% (395,918), validation (20,121) and test (40,152), where the number of
% annotated images in each set is in the parentheses. 
%validation and test. 
The training set is
not exhaustively annotated because of its sheer size.
%due to its large number of images. 
There are also fewer annotated objects per training image than the validation and
testing image (on average 1.53 objects for training {\it vs.} 2.5
objects for validation set). We follow all the experiment settings
as in \cite{hoffman2014lsda},
and simulate having access to image-level annotations for all 200 categories
and bounding box annotations only for the first 100 categories (alphabetical
order). We separate the dataset into classification and detection
sets. For the classification data, we use 200,000 images in total from
all 200 categories of the training subset (around 1,000 images per
category) and their image-level labels. The validation set is roughly
split in half: val1 and val2 as in \cite{girshick2014rcnn}. For the
detection training set, we take the images with their bounding boxes
from only the first 100 categories ($\mathcal{B}$) in val1 (around 5,000
images in total). Since the validation dataset is relatively small,
we then augment val1 with 1,000 bounding box annotated images per class
from the training set (following the same protocol of
\cite{girshick2014rcnn, hoffman2014lsda}). Finally, we evaluate
our knowledge transfer framework on the val2 dataset (9,917 images in total).

%-------------------------------------------------------------------------
\subsection{Implementation Details}
In all the experiments, we consider LSDA \cite{hoffman2014lsda} as our
baseline model and follow their main settings. 
Following \cite{hoffman2014lsda}, we first use the Caffe
\cite{jia2014caffe} implementation of the ``\emph{AlexNet}'' CNN. 
A pre-trained CNN on ILSVRC 2012 dataset is then fine-tuned
on the classification training dataset (see Section \ref{dataset}).
This CNN is then fine-tuned again for detection on the labeled region
proposals of the first 100 categories (subset $\mathcal{B}$) of val1.
Selective Search \cite{UijlingsIJCV13} with ``fast'' mode is adopted
to generate the region proposals from all the images in val1 and val2.
We also report results using two deeper models of ``\emph{VGG-Nets}'' \cite{Simonyan14c}, namely, the 16-layer model (\emph{VGG-16}) and the 19-layer model (\emph{VGG-19}), \emph{GoogLeNet}\cite{Szegedy_2016_CVPR} and two \emph{ResNets}\cite{He_2016_CVPR} (34-layer and 50-layer) with the Caffe toolbox.
For the semantic representation, we use word2vec CBoW embeddings pre-trained on part of the Google News dataset comprising about 100 billion words~\cite{Mikolov:13}. We train AutoExtend~\cite{Rothe:15} using WordNet 3.0 to obtain synset embeddings, and using equal weights for the synset, lexeme and WordNet relation constraints ($\alpha = \beta = 0.33$). As all categories are nouns, we use only hypernyms as the WordNet relation constraint.
For the sparse representation of a target word vector in Eq.\ \eqref{recon_func}, we limit the maximum number of non-zero components to 20, since a target category has strong correlation with a small number of source categories. We set $\lambda = 100$ in Eq.\ \eqref{recon_func} and $\lambda_0 = 1000$ in Eq.\ \eqref{least_sq_func} based on a validation set. 
Other detailed information regarding training and detection can be found in Section \ref{background}.
%%%%%%%%%%%%%%%%Fig results%%%%%%%%%%%%%%%%%%%%%%%%%%%%%%%%%%%%%%%%%%
\begin{figure*}[htbp!] % Fig results
  \centering
  \includegraphics[width=\linewidth]{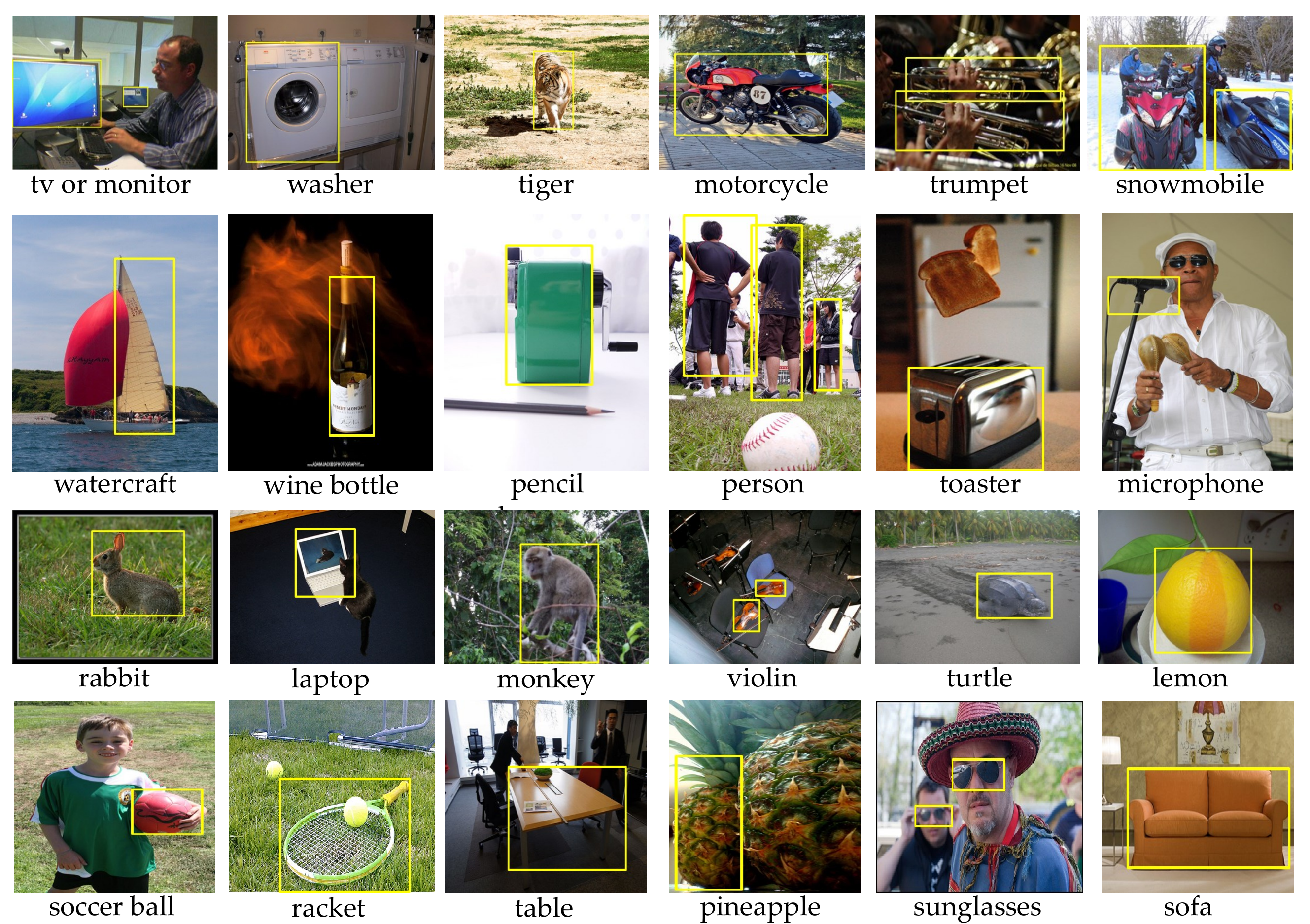}
  \caption{Examples of correct detections (true positives) of our mixture knowledge transfer model on ILSVRC2013 images. For each image, only detections for the ``weakly labeled'' target category (text below image) are listed.}
  \label{fig:results_tp}
  % \vspace{2mm}
\end{figure*}
%%%%%%%%%%%%%%%%Fig%%%%%%%%%%%%%%%%%%%%%%%%%%%%%%%%%%%%%%%%%%

%%%%%%%%%%%%%%%%Fig results False positive%%%%%%%%%%%%%%%%%%%%%%%%%%%%%%%%%%%%%%%%%%
\begin{figure*}[htbp!] % Fig results
  \centering
  \includegraphics[width=\linewidth]{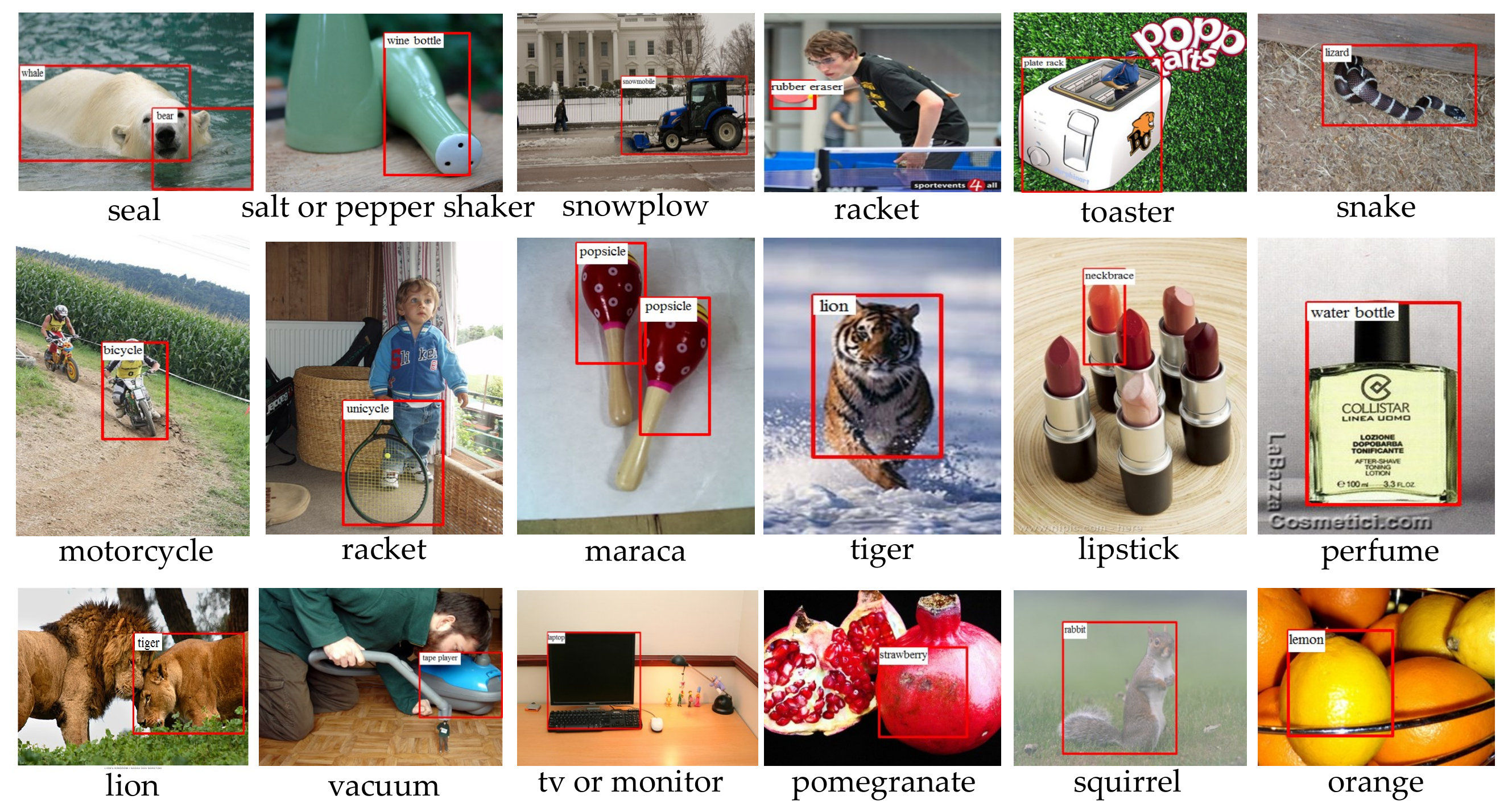}
  \caption{Examples of incorrect detections (confusion with other objects) of our mixture knowledge transfer model on ILSVRC2013 images. The detected object label is shown in the top-left of its bounding box.}
  \label{fig:results_fp}
  %\vspace{-3mm}
\end{figure*}
%%%%%%%%%%%%%%%%%Fig%%%%%%%%%%%%%%%%%%%%%%%%%%%%%%%%%%%%%%%%%%

%-------------------------------------------------------------------------
\subsection{Quantitative Evaluation on the ``Weakly Labeled'' Categories with ``\emph{AlexNet}''}
\label{results_eval}
Setting LSDA as the baseline, we compare the detection performance
of our proposed knowledge transfer methods against LSDA. The results are
summarized in Table \ref{table-results}. As we are concerned with the detection of
the ``weakly labeled'' categories, we focus mainly on the second column of the
table (mean average precision (mAP) on $\mathcal{A}$). Rows 1-5 in Table
\ref{table-results} are the baseline results for LSDA. The first row shows the
detection results by applying a classification network ({\it i.e.}, weakly
supervised learning, and without adaptation) trained with only classification
data, achieving only an mAP of 10.31\% on the ``weakly labeled'' 100 categories.
The last row shows the results of an oracle detection network which
assumes that bounding boxes for all 200 categories are available ({\it i.e.},
supervised learning). This is treated as the upper bound (26.25\%) of the
fully supervised framework.
We observed that the best result obtained by LSDA is to adapt both category
independent and category specific layers, %(namely, to adapt the feature
%representation and category-specific parameters together)
and transforming with
the weighted $fc8$ layer weight change of its 100 nearest neighbor categories (\textbf{weighted-100} with 16.33\% in Table \ref{table-results}). Our ``\textit{weighted}'' scheme works consistently better than its ``\textit{average}'' counterpart.
% Note: nearest neighbors
% are defined in the fourth step of Section \ref{background}, namely categories
% with the smallest Euclidean distance of the $fc8$ weights to the target category).

%%%%%%%%%%%%%%%%%%%%%%%%%%%%%%%table%%%%%%%%%%%%%%%%%%%%%%%%%%%%%%%%%%%%%%%%%%%%%%
\begin{table}[t] %\fontsize{9pt}{9pt}\selectfont
\renewcommand{\arraystretch}{1.4}
\caption{Comparison of mean average precision (mAP) for semantic similarity measures/representations, using \textbf{Weighted - 100}.} 
\label{table-results-semantic}
\centering
\begin{tabular}{l|c c c c}
  \hline
  Method & \tabincell{c}{Path\\ Similarity} & \tabincell{c}{Lin \\ Similarity } & \tabincell{c}{Naive\\ Embeddings}  & \tabincell{c}{AutoExtend\\(this paper)}\\ \hline
   mAP &17.08 & 17.31 & 17.83 & 18.32\\ \hline
\end{tabular}
\end{table}
%%%%%%%%%%%%%%%%%%%%%%%%%%%%%%%%%%%%%%%%%%%%%%%%%%%%%%%%%%%%%%%%%%%%%%%%%%%%%%%%%%%%%

For our \textbf{visual knowledge transfer model}, we show steady improvement over the
baseline LSDA methods when considering the average weight change of both 5 and 10
visually similar categories, with 1.45\% and 1.47\% increase in mAP, respectively.
This proves that our proposed visual similarity measure %strategy of finding visually similar categories 
is superior
to that of LSDA, showing that category-specific adaptation can indeed be improved 
%it is effective to convert a classifier into a detector based on the knowledge of its visually similar categories. 
based on knowledge about the visual similarities between categories. Further improvement is
achieved by modeling individual weights of all 100 source categories according
to their degree of visual similarities to the target category (\textbf{weighted-100}
with 19.02\% in the table). This verifies our supposition that the transformation
from a classifier to a detector of a certain category is more related to
visually similar categories, and is proportional to their degrees of similarity.
For example, \emph{motorcycle} is most similar to \emph{bicycle}.
Thus the weight change from a \emph{bicycle} classifier to detector has
the largest influence on the transformation of \emph{motorcycle}. The influence of 
less visually relevant categories, such as \emph{cart} and \emph{chain saw},
is much smaller. For visually dissimilar categories (\emph{apple, fig, hotdog}, etc.), the influence is extremely insignificant.
We show some examples of visual similarities between a target category and
its source categories in the first row of Fig.~\ref{fig:relation}. For each
target category, the top-10 weighted nearest neighbor categories with their similarity
degrees are visualized.

Our \textbf{semantic knowledge transfer model} also showed marked improvement
over the LSDA baseline (Table \ref{table-results}, Rows 9-12), and is
comparable to the results of the visual transfer model. This suggests that the cross-domain knowledge transfer from semantic relatedness to visual similarity is
very effective. The best performance for the semantic transfer model (19.04\%) is
obtained by sparsely reconstructing the target category with the source categories
using the synset embeddings. 
We also compare the results of using other semantic similarity measures, as shown in Table \ref{table-results-semantic}.
The result of using synset embeddings (18.32\%, using weighted-100, the same below) are superior to using `naive' word2vec embeddings (17.83\%) and WordNet based measures such as path-based similarity (17.08\%) and  Lin similarity~\cite{Lin:98} (17.31\%). %Using relative distance based on the WordNet tree structure produces inferior results: \ie, path-based similarity with 17.08\%, Lin similarity~\cite{Lin:98} with 17.31\%, and `naive' word2vec embeddings with 17.83\%. 
Several examples visualizing the related categories
of the 10 largest semantic reconstruction coefficients are shown in the middle row
of Fig.~\ref{fig:relation}. We observe that semantic relatedness indeed correlates
with visual similarity.% of object categories.

\begin{figure}[b!]
  \centering
  \includegraphics[width=0.85\columnwidth]{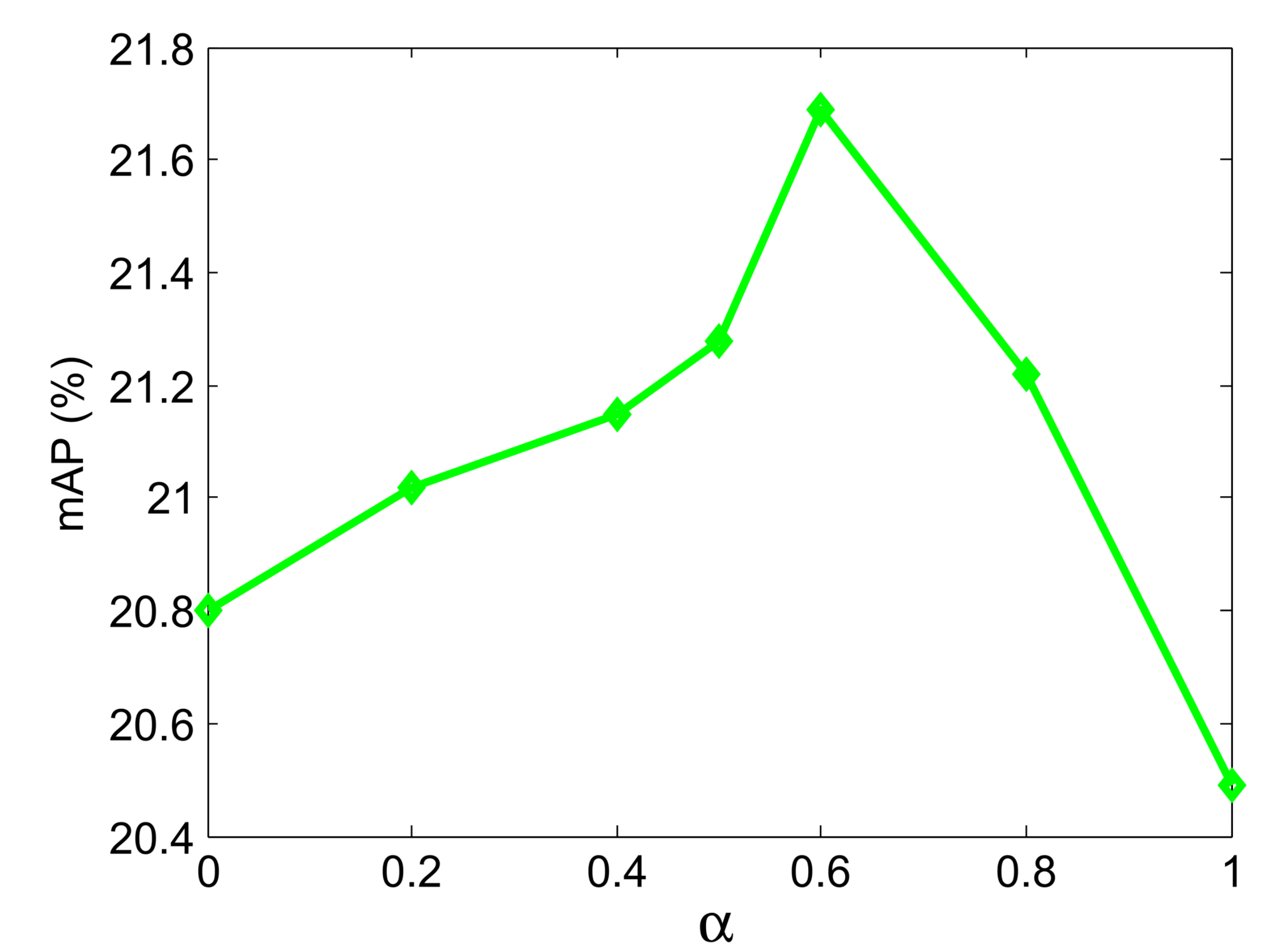}
  \caption{Sensitivity of parameter $\alpha$ vs. mAP for detection of ``weakly labeled'' categories on the validation (val1) dataset. $\alpha \in [0,1]$ is a parameter used to control the relative influence of the two similarity measurements. $\alpha$ is set to 1 when only considering visual similarity transfer, and 0 for semantic similarity transfer.}
% We find that best validation performance is obtained by setting $\alpha$ to 0.6 on the val1 dataset.}
  \label{fig:alpha}
\end{figure}

The state-of-the-art result using the 8-layer ``\emph{Alex-Net}'' for semi-supervised detection on this dataset is
achieved by our\textbf{ mixture transfer model} which combines visual similarity
and semantic relatedness. A boost in performance of 3.88\% on original split (3.82\%$\pm$0.12\%, based on 6 different splits of the dataset) is achieved over
the best result reported by LSDA %(20.03\% {\it vs.}\ 16.15\%) 
on the ``weakly
labeled'' categories. We show examples of transferred categories with
their corresponding weights for several target categories in the bottom row of
Fig.~\ref{fig:relation}. The parameter $\alpha$ in Eq.\ \eqref{eq_fusion} for the
mixture model weights is set to 0.6 for final detection, where $\alpha \in
\{0, 0.2, 0.4, 0.5, 0.6, 0.8, 1\}$ is chosen via cross-validation on the
val1 detection set (Fig.~\ref{fig:alpha}). 
This suggests that the transferring
of visual similarities is slightly more important than semantic relatedness, although
both are indeed complementary. 
We do not tune $\alpha$ for each category separately, though this can be expected to further improve our detection performance.
Fig.~\ref{fig:results_tp} and Fig.~\ref{fig:results_fp} show some examples of correct and incorrect detections %on the ``weakly labeled'' categories 
respectively. Although our proposed mixture
transfer model achieves the state of the art in detecting the ``weakly
labeled'' categories, it is still occasionally confused by %occasionally error-prone in confusing 
visually similar categories.

%%%%%%%%%%%%%%%%Fig alpha%%%%%%%%%%%%%%%%%%%%%%%%%%%%%%%%%%%%%%%%%%
%%%%%%%%%%%%%%%%Fig%%%%%%%%%%%%%%%%%%%%%%%%%%%%%%%%%%%%%%%%%%

%------------------------------------------------------------------------
\subsection{Experimental Results with Deeper Neural Networks}
\label{results_vgg}

%%%%%%%%%%%%%%%%%%%%%%%%%%%%%%%table%%%%%%%%%%%%%%%%%%%%%%%%%%%%%%%%%%%%%%%%%%%%%%
\begin{table*}[!t] \fontsize{9pt}{9pt}\selectfont
\renewcommand{\arraystretch}{1.4}
\caption{Comparisons of detection mean average precision (mAP) on the ``weakly labeled'' categories of ILSVRC2013 val2, using ``\textbf{\textit{VGG-Nets}}'', ``\textbf{\textit{GoogLeNet}}'' and ``\textbf{\textit{ResNets}}''. For LSDA, our visual similarity and semantic relatedness transfer models, \textbf{Weighted - 100} scheme is adopted.} 
\label{table-results-vgg}
\centering
\begin{tabular}{l| c c c c c c c}
  \hline
   Method & \tabincell{c}{Only\\ classification} & \tabincell{c}{LSDA \\ class invariant } & \tabincell{c}{LSDA\\ class invariant \& specific}  & \tabincell{c}{\textbf{Ours}\\\textbf{visual}}  & \tabincell{c}{\textbf{Ours}\\\textbf{semantic}}  & \tabincell{c}{\textbf{Ours}\\\textbf{mixed}} & \tabincell{c}{\textbf{Ours}\\\textbf{mixed + BB reg.}}\\ \hline
   \textit{Alex-Net} &10.31 & 15.85 & 16.33 & 19.02 &18.32 &20.03 &21.88\\ \hline
   \textit{VGG-16} &14.89 & 18.24 & 18.86 & 21.75 &21.07 &23.21 &24.91\\ \hline
   \textit{VGG-19} &16.22 & 20.38 & 21.02 & 23.89 &23.10 &25.07 &27.32\\ \hline
    \textit{GoogLeNet} &16.12 & -- & -- & 23.62 &23.46 &25.06 &26.62\\ \hline
   \textit{ResNet-34} &17.09 & -- & -- & 24.28 &24.64 &26.17 &28.05\\ \hline
   \textit{ResNet-50} &17.34 & -- & -- & 24.94 &24.75 &26.77 &28.30\\ \hline
\end{tabular}
\end{table*}
%%%%%%%%%%%%%%%%%%%%%%%%%%%%%%%%%%%%%%%%%%%%%%%%%%%%%%%%%%%%%%%%%%%%%%%%%%%%%%%%%%%%%
Previous work \cite{Simonyan14c, rbg_rcnn_pami, He_2016_CVPR} found that region based CNN detection performance is significantly influenced by the choice of CNN architecture. In Table \ref{table-results-vgg}, we show some detection results using the 16-layer and 19-layer deep ``\textit{VGG-Nets}'' proposed by Simonyan and Zisserman \cite{Simonyan14c}, ``\textit{GoogLeNet}'' (Inception-v2\cite{Szegedy_2016_CVPR}), together with the 34-layer and 50-layer ``\textit{ResNets}'' \cite{He_2016_CVPR}. The \textit{VGG-16} network consists of 13 convolutional layers of very small ($3\times3$) convolution filters, with 5 max pooling layers interspersed, and topped with 3 fully connected layers (namely, $fc6$, $fc7$ and $fc8$). The \textit{VGG-19} network extends \textit{VGG-16} by inserting 3 more convolutional layers, while keeping other layer configurations unchanged. The state-of-the-art residual networks (\textit{ResNets}) make use of identity shortcut connections that enable flow of information across layers without decay. In the aforementioned deep neural nets, we transfer the parameters of the last fully connected layer ($fc8$ layer for \textit{VGG-Nets} and the only $fc$ layer in \textit{GoogLeNet} and \textit{ResNets}).

As can be seen from Table \ref{table-results-vgg}, the very deep ConvNets \textit{VGG-16}, \textit{VGG-19} and \textit{GoogLeNet} significantly outperform  \textit{Alex-Net} for all the adaptation methods. Our knowledge transfer models using the very deep \textit{VGG-Nets} with different similarity measures show consistent improvement over the LSDA baseline method. The relative overall improvement over performance using \textit{VGG-Nets} is similar with that of \textit{AlexNet}. \textit{GoogLeNet} obtains similar results to \textit{VGG-19}, while the best performance is achieved by \textit{ResNet-50}. 
% In principle, we would expect further improvement by using the \textit{Res-Net} \cite{He_2016_CVPR} which has more than 150 layers. However, testing various deeper networks is beyond the scope of this paper.

\subsection{Experimental Results with Bounding-box Regression}
\label{results_reg}

%%%%%%%%%%%%%%%%Fig bb%%%%%%%%%%%%%%%%%%%%%%%%%%%%%
\begin{figure*}[h] % Fig bb
  \centering
  \includegraphics[width=\linewidth]{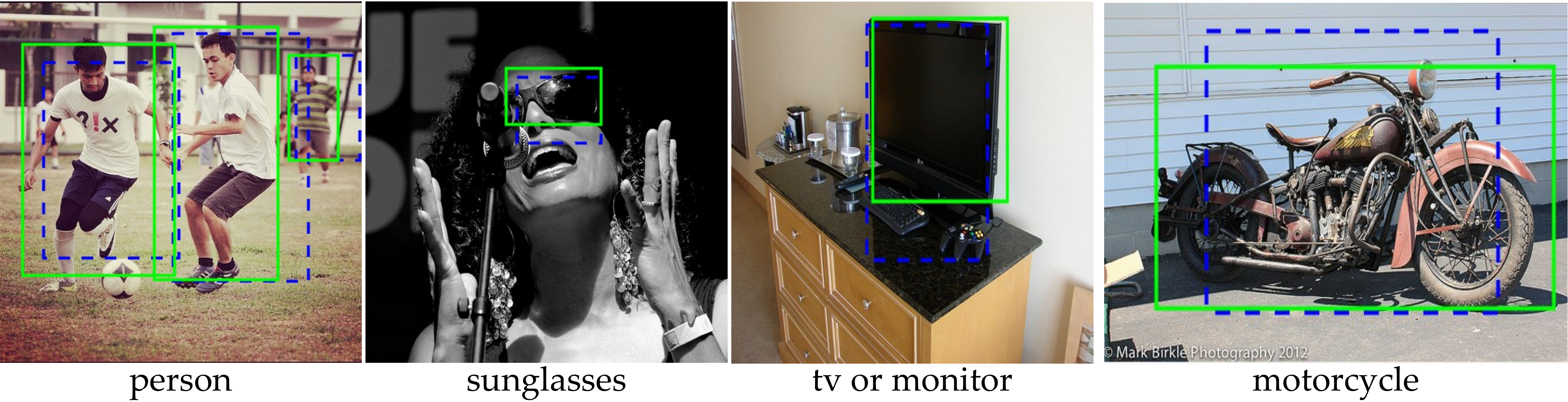}
  \caption{Some example detections before and after bounding box regression on the ``weakly labeled'' categories. Boxes before (resp. after) bounding-box regression are shown in dashed blue (resp. green).}
  \label{fig:bb}
\end{figure*}
%%%%%%%%%%%%%%%%%Fig%%%%%%%%%%%%%%%%%%%%%%%%%%%%%%%%%%%%%%%%%%

The results in Table \ref{table-results} and Table \ref{table-results-vgg} show that the transferred bounding-box regression from ``fully labeled'' categories fixes a large number of incorrectly localized detections%resulted by mis-localization
, boosting mAP by about 2 points for the ``weakly labeled'' categories. The bounding-box regression process could boost mAP by 3 to 4 points if the bounding box annotations for all the categories were provided. We show some example detections before and after bounding box regression on the ``weakly labeled'' categories in Fig.~\ref{fig:bb}, using \textit{VGG-16}. 

In addition to the results reported above using the default Intersection-over-Union (IoU) threshold $0.5$, we evaluate detection performance by setting different IoU overlap ratios $\{0.6, 0.7\}$ before and after bounding box regression using the best performing network, \ie, \textit{ResNet-50}. The mAP@IoU=0.6 is 23.21 \textit{v.s.} 26.02 before and after regression, mAP@IoU=0.7 is 17.89 \textit{v.s.} 20.64, respectively. These results validate that the proposed regression transfer is very effective in moving bounding box boundaries so as to cover more foreground object regions, by transferring the class-specific regressors from ``fully labeled'' categories to ``weakly labeled'' categories based on the proposed similarity measures. Note that results in other parts of this paper are reported as mAP@IoU=0.5, unless specified otherwise.

\subsection{Experimental Results with Fast R-CNN}
\label{fast-rcnn}
The proposed knowledge transfer method is applied to Fast R-CNN\cite{girshick15fastrcnn}, without modifying much of the framework. Our Fast R-CNN based transfer framework is much faster than the R-CNN based approach, since in Fast R-CNN, in Fast R-CNN, an image is first fed into a CNN to create a convolution feature map and a single feature vector is then extracted from a Region of Interest (RoI) pooling layer for each region proposal, while in R-CNN, each region proposal in an image is fed into a CNN to extract feature independently, which is considerably more computationally expensive than Fast R-CNN.

We investigate two different bounding box regression strategies in Fast R-CNN. For the first strategy, we remove the built-in bounding-box regression layer in the Fast R-CNN pipeline and transfer the regressor off-line after detection as in Section \ref{bb_reg}, like the \textit{SPP-Net}\cite{SPPnet_TPAMI2015}, which we call ``2-stage'' Fast R-CNN. For the second strategy, we use the built-in bounding-box regression layer, which is actually a class-specific fully-connected layer with $4C$ neurons (4 indicates the coordinates of a bounding box position, $C$ indicates the number of categories). This class-specific layer can be therefore transferred from ``fully labeled'' categories to ``weakly labeled'' categories using the proposed similarity measures in a  way similar to that described in Section \ref{bb_reg}. This strategy is called ``end-to-end''. Table \ref{table-fast-rcnn} shows the detection performance using these Fast R-CNN based ConvNets. As can be seen from the table, Fast R-CNN achieves consistently better performance over the R-CNN based approach, and the ``end-to-end'' joint training/testing is superior to the ``2-stage'' pipeline. State-of-the-art detection performance on the weakly labeled categories (29.71) is obtained by Fast R-CNN based \textit{ResNet-50}, which is very close to that of the fully supervised R-CNN based \textit{AlexNet} (30.82).
%%%%%%%%%%%%%%%%%%%%%%%%%%%%%%%table%%%%%%%%%%%%%%%%%%%%%%%%%%%%%%%%%%%%%%%%%%%%%%
\begin{table}[t] %\fontsize{9pt}{9pt}\selectfont
\renewcommand{\arraystretch}{1.4}
\caption{Detection performance using Fast R-CNN on the ``weakly labeled'' categories of ILSVRC2013 val2.} 
\label{table-fast-rcnn}
\centering
\begin{tabular}{l|c c c}
  \hline
  Method & \tabincell{c}{\textit{Alex-Net}} & \tabincell{c}{\textit{VGG-16}}  & \tabincell{c}{\textit{ResNet-50}}\\ \hline
   2-stage &22.36 & 25.47 & 29.09\\ \hline
   end-to-end &22.79 & 26.22 & 29.71\\ \hline
\end{tabular}
\end{table}
%%%%%%%%%%%%%%%%%%%%%%%%%%%%%%%%%%%%%%%%%%%%%%%%%%%%%%%%%%%%%%%%%%%%%%%%%%%%%%%%%%%%%
%------------------------------------------------------------------------
\section{Conclusion}
\label{semi-concl}
In this paper, we investigated how knowledge about object similarities from both visual and semantic domains can be transferred to adapt an image classifier to an object detector in a semi-supervised setting. We experimented with different CNN architectures, found clear evidence that both visual and semantic similarities play an essential role in improving the adaptation process, and that the combination of the two modalities yielded state-of-the-art performance, suggesting that knowledge inherent in visual and semantic domains is complementary. Future work includes extracting more knowledge from different domains, using better representations, and investigating the possibility of using category-invariant properties, \eg, the difference between feature distributions of whole images and target objects, to help knowledge transfer. We believe that the combination of knowledge from different domains is key to improving semi-supervised object detection. 
% or


% \appendices
% \section{1st}
% This appendix provides details about the 

% \section{}

\ifCLASSOPTIONcompsoc
  \section*{Acknowledgments}
\else
  \section*{Acknowledgment}
\fi

%This work was partially supported by French Research Agency, Agence Nationale de Recherche (ANR), by the VideoSense Project under Grant 2009 CORD 026 02, and by the Visen project, under grant ANR-12-CHRI-0002-04, within the framework of the ERA-Net CHIST-ERA. The authors thank NVIDIA for providing the Titan X GPU.
This work was partly supported by the French Research Agency, Agence Nationale de Recherche (ANR),  through the VideoSense Project under Grant 2009 CORD 026 02, and the Visen project under Grants ANR-12-CHRI-0002-04 and UK EPSRC EP/K019082/1 within the framework of the ERA-Net CHIST-ERA, and by the Partner University Foundation through the 4D Vision project. The authors thank NVIDIA for providing the Titan X GPU.

\ifCLASSOPTIONcaptionsoff
  \newpage
\fi



%\bibliographystyle{IEEEtran}


% \end{thebibliography}

\bibliographystyle{IEEEtran}
\bibliography{IEEEabrv,reference}

\begin{thebibliography}{10}
\providecommand{\url}[1]{#1}
\csname url@samestyle\endcsname
\providecommand{\newblock}{\relax}
\providecommand{\bibinfo}[2]{#2}
\providecommand{\BIBentrySTDinterwordspacing}{\spaceskip=0pt\relax}
\providecommand{\BIBentryALTinterwordstretchfactor}{4}
\providecommand{\BIBentryALTinterwordspacing}{\spaceskip=\fontdimen2\font plus
\BIBentryALTinterwordstretchfactor\fontdimen3\font minus
  \fontdimen4\font\relax}
\providecommand{\BIBforeignlanguage}[2]{{%
\expandafter\ifx\csname l@#1\endcsname\relax
\typeout{** WARNING: IEEEtran.bst: No hyphenation pattern has been}%
\typeout{** loaded for the language `#1'. Using the pattern for}%
\typeout{** the default language instead.}%
\else
\language=\csname l@#1\endcsname
\fi
#2}}
\providecommand{\BIBdecl}{\relax}
\BIBdecl

\bibitem{AlexNet_NIPS2012}
A.~Krizhevsky, I.~Sutskever, and G.~E. Hinton, ``Imagenet classification with
  deep convolutional neural networks,'' in \emph{Neural Information Processing
  Systems ({NIPS})}, 2012.

\bibitem{Szegedy_NIPS2013}
C.~Szegedy, A.~Toshev, and D.~Erhan, ``Deep neural networks for object
  detection,'' in \emph{Neural Information Processing Systems ({NIPS})}, 2013,
  pp. 2553--2561.

\bibitem{sermanet-iclr-14}
P.~Sermanet, D.~Eigen, X.~Zhang, M.~Mathieu, R.~Fergus, and Y.~LeCun,
  ``Overfeat: Integrated recognition, localization and detection using
  convolutional networks,'' in \emph{International Conference on Learning
  Representations (ICLR)}, 2014.

\bibitem{girshick2014rcnn}
R.~Girshick, J.~Donahue, T.~Darrell, and J.~Malik, ``Rich feature hierarchies
  for accurate object detection and semantic segmentation,'' in \emph{IEEE
  Conference on Computer Vision and Pattern Recognition ({CVPR})}, 2014.

\bibitem{SPPnet_TPAMI2015}
K.~He, X.~Zhang, S.~Ren, and J.~Sun, ``Spatial pyramid pooling in deep
  convolutional networks for visual recognition,'' \emph{IEEE Transactions on
  Pattern Analysis and Machine Intelligence (TPAMI)}, vol.~37, no.~9, pp.
  1904--1916, 2015.

\bibitem{girshick15fastrcnn}
R.~Girshick, ``Fast {R-CNN:},'' in \emph{International Conference on Computer
  Vision ({ICCV})}, 2015, pp. 1440--1448.

\bibitem{ren15fasterrcnn}
S.~Ren, K.~He, R.~Girshick, and J.~Sun, ``Faster r-cnn: Towards real-time
  object detection with region proposal networks,'' in \emph{Neural Information
  Processing Systems ({NIPS})}, 2015, pp. 91--99.

\bibitem{Redmon_2016_CVPR}
J.~Redmon, S.~Divvala, R.~Girshick, and A.~Farhadi, ``You only look once:
  Unified, real-time object detection,'' in \emph{IEEE Conference on Computer
  Vision and Pattern Recognition (CVPR)}, 2016.

\bibitem{liu2016single}
W.~{Liu}, D.~{Anguelov}, D.~{Erhan}, C.~{Szegedy}, S.~{Reed}, C.~{Fu}, and
  A.~{Berg}, ``{SSD}: Single shot multibox detector,'' in \emph{European
  Conference on Computer Vision (ECCV)}, 2016, pp. 21--37.

\bibitem{Everingham-et-al-IJCV10}
M.~Everingham, L.~Van~Gool, C.~Williams, J.~Winn, and A.~Zisserman, ``The
  {Pascal} visual object classes ({VOC}) challenge,'' \emph{International
  Journal of Computer Vision (IJCV)}, vol.~88, no.~2, pp. 303--338, 2010.

\bibitem{ILSVRC15}
O.~Russakovsky, J.~Deng, H.~Su, J.~Krause, S.~Satheesh, S.~Ma, Z.~Huang,
  A.~Karpathy, A.~Khosla, M.~Bernstein, A.~C. Berg, and L.~Fei-Fei, ``{ImageNet
  Large Scale Visual Recognition Challenge},'' \emph{International Journal of
  Computer Vision (IJCV)}, vol.~0, no.~0, pp. 1--42, April 2015.

\bibitem{MSCOCO14}
T.~Lin, M.~Maire, S.~Belongie, L.~D. Bourdev, R.~B. Girshick, J.~Hays,
  P.~Perona, D.~Ramanan, P.~Doll{\'{a}}r, and C.~L. Zitnick, ``Microsoft
  {COCO:} common objects in context,'' in \emph{European Conference on Computer
  Vision (ECCV)}, 2014, pp. 740--755.

\bibitem{hoffman2014lsda}
J.~Hoffman, S.~Guadarrama, E.~Tzeng, R.~Hu, J.~Donahue, R.~Girshick,
  T.~Darrell, and K.~Saenko, ``{LSDA}: Large scale detection through
  adaptation,'' in \emph{Neural Information Processing Systems ({NIPS})}, 2014.

\bibitem{zhou2014object}
B.~Zhou, A.~Khosla, A.~Lapedriza, A.~Oliva, and A.~Torralba, ``Object detectors
  emerge in deep scene {CNNs},'' in \emph{International Conference on Learning
  Representations (ICLR)}, 2015.

\bibitem{Deselaers_sim10CVPR}
T.~Deselaers and V.~Ferrari, ``Visual and semantic similarity in imagenet,'' in
  \emph{IEEE Conference on Computer Vision and Pattern Recognition ({CVPR})},
  2011.

\bibitem{Tang_2016_CVPR}
Y.~Tang, J.~Wang, B.~Gao, E.~Dellandrea, R.~Gaizauskas, and L.~Chen, ``Large
  scale semi-supervised object detection using visual and semantic knowledge
  transfer,'' in \emph{IEEE Conference on Computer Vision and Pattern
  Recognition (CVPR)}, 2016.

\bibitem{Crandall-et-al-ECCV06}
D.~Crandall and D.~Huttenlocher, ``Weakly supervised learning of part-based
  spatial models for visual object recognition,'' in \emph{European Conference
  on Computer Vision (ECCV)}, 2006.

\bibitem{Chum07a}
O.~Chum and A.~Zisserman, ``An exemplar model for learning object classes,'' in
  \emph{2007 IEEE Conference on Computer Vision and Pattern Recognition}, 2007,
  pp. 1--8.

\bibitem{Galleguillos_ECCV08}
C.~Galleguillos, B.~Babenko, A.~Rabinovich, and S.~Belongie, ``Weakly
  supervised object recognition and localization with stable segmentations,''
  in \emph{European Conference on Computer Vision (ECCV)}, 2008.

\bibitem{Nguyen-et-al-ICCV09}
M.~Nguyen, L.~Torresani, F.~{de la Torre}, and C.~Rother, ``Weakly supervised
  discriminative localization and classification: a joint learning process,''
  in \emph{International Conference on Computer Vision (ICCV)}, 2009, pp.
  1925--1932.

\bibitem{Siva-et-al-ICCV11}
P.~Siva and T.~Xiang, ``Weakly supervised object detector learning with model
  drift detection,'' in \emph{International Conference on Computer Vision
  (ICCV)}, 2011.

\bibitem{Pandey-et-al-ICCV11}
M.~Pandey and S.~Lazebnik, ``Scene recognition and weakly supervised object
  localization with deformable part-based models,'' in \emph{International
  Conference on Computer Vision (ICCV)}, 2011.

\bibitem{siva2012weakly}
P.~Siva, C.~Russell, and T.~Xiang, ``In defence of negative mining for
  annotating weakly labelled data,'' in \emph{European Conference on Computer
  Vision (ECCV)}, 2012.

\bibitem{Deselaers-et-al-IJCV12}
T.~Deselaers, B.~Alexe, and V.~Ferrari, ``Weakly supervised localization and
  learning with generic knowledge,'' \emph{International Journal of Computer
  Vision (IJCV)}, vol. 100, no.~3, pp. 275--293, 2012.

\bibitem{Shi_2013_ICCV}
Z.~Shi, T.~M. Hospedales, and T.~Xiang, ``Bayesian joint topic modelling for
  weakly supervised object localisation,'' in \emph{International Conference on
  Computer Vision (ICCV)}, 2013, pp. 2984--2991.

\bibitem{ytangICIP2014}
Y.~Tang, X.~Wang, E.~Dellandrea, S.~Masnou, and L.~Chen, ``Fusing generic
  objectness and deformable part-based models for weakly supervised object
  detection,'' in \emph{IEEE International Conference on Image Processing
  (ICIP)}, 2014.

\bibitem{bilen2014bmvc}
H.~Bilen, M.~Pedersoli, and T.~Tuytelaars, ``Weakly supervised object detection
  with posterior regularization,'' in \emph{British Machine Vision Conference
  ({BMVC})}, 2014.

\bibitem{songICML14slsvm}
H.~O. Song, R.~Girshick, S.~Jegelka, J.~Mairal, Z.~Harchaoui, and T.~Darrell,
  ``On learning to localize objects with minimal supervision,'' in
  \emph{International Conference on Machine Learning ({ICML})}, 2014.

\bibitem{Bilen_2015_CVPR}
H.~Bilen, M.~Pedersoli, and T.~Tuytelaars, ``Weakly supervised object detection
  with convex clustering,'' in \emph{IEEE Conference on Computer Vision and
  Pattern Recognition ({CVPR})}, 2015.

\bibitem{CWang_LCL_TIP2015}
C.~Wang, K.~Huang, W.~Ren, J.~Zhang, and S.~Maybank, ``Large-scale weakly
  supervised object localization via latent category learning,'' \emph{IEEE
  Transactions on Image Processing (TIP)}, vol.~24, no.~4, pp. 1371--1385,
  April 2015.

\bibitem{Tang_TMM_WSL}
Y.~Tang, X.~Wang, E.~Dellandrea, and L.~Chen, ``Weakly supervised learning of
  deformable part-based models for object detection via region proposals,''
  \emph{IEEE Transactions on Multimedia (TMM)}, vol.~PP, no.~99, pp. 1--1,
  2016.

\bibitem{UijlingsIJCV13}
J.~Uijlings, K.~van~de Sande, T.~Gevers, and A.~Smeulders, ``Selective search
  for object recognition,'' \emph{International Journal of Computer Vision
  (IJCV)}, vol. 104, no.~2, pp. 154--171, 2013.

\bibitem{cheng2014bing}
M.-M. Cheng, Z.~Zhang, W.-Y. Lin, and P.~Torr, ``Bing: Binarized normed
  gradients for objectness estimation at 300fps,'' in \emph{IEEE Conference on
  Computer Vision and Pattern Recognition (CVPR)}, 2014.

\bibitem{ZitnickECCV14edgeBoxes}
C.~L. Zitnick and P.~Doll\'ar, ``Edge boxes: Locating object proposals from
  edges,'' in \emph{European Conference on Computer Vision ({ECCV})}, 2014.

\bibitem{Oquab_2015_CVPR}
M.~Oquab, L.~Bottou, I.~Laptev, and J.~Sivic, ``Is object localization for
  free? - weakly-supervised learning with convolutional neural networks,'' in
  \emph{IEEE Conference on Computer Vision and Pattern Recognition (CVPR)},
  2015.

\bibitem{Bilen16}
H.~Bilen and A.~Vedaldi, ``Weakly supervised deep detection networks,'' in
  \emph{IEEE Conference on Computer Vision and Pattern Recognition (CVPR)},
  2016.

\bibitem{zhou2016cvpr}
B.~Zhou, A.~Khosla, L.~A., A.~Oliva, and A.~Torralba, ``{Learning Deep Features
  for Discriminative Localization.}'' \emph{IEEE Conference on Computer Vision
  and Pattern Recognition (CVPR)}, 2016.

\bibitem{Shao_TL_survey}
L.~Shao, F.~Zhu, and X.~Li, ``Transfer learning for visual categorization: A
  survey,'' \emph{IEEE Transactions on Neural Networks and Learning Systems
  (TNNLS)}, vol.~26, no.~5, pp. 1019--1034, May 2015.

\bibitem{Donahue_CVPR2013}
J.~Donahue, J.~Hoffman, E.~Rodner, K.~Saenko, and T.~Darrell, ``Semi-supervised
  domain adaptation with instance constraints,'' in \emph{IEEE Conference on
  Computer Vision and Pattern Recognition (CVPR)}, 2013.

\bibitem{Oquab14}
M.~Oquab, L.~Bottou, I.~Laptev, and J.~Sivic, ``Learning and transferring
  mid-level image representations using convolutional neural networks,'' in
  \emph{IEEE Conference on Computer Vision and Pattern Recognition (CVPR)},
  2014.

\bibitem{Rochan_2015_CVPR}
M.~Rochan and Y.~Wang, ``Weakly supervised localization of novel objects using
  appearance transfer,'' in \emph{IEEE Conference on Computer Vision and
  Pattern Recognition ({CVPR})}, 2015.

\bibitem{shu_DTNs_MM}
X.~Shu, G.-J. Qi, J.~Tang, and J.~Wang, ``Weakly-shared deep transfer networks
  for heterogeneous-domain knowledge propagation,'' in \emph{ACM International
  Conference on Multimedia (MM)}, 2015.

\bibitem{zhu2011heterogeneous}
Y.~Zhu, Y.~Chen, Z.~Lu, S.~J. Pan, G.-R. Xue, Y.~Yu, and Q.~Yang,
  ``Heterogeneous transfer learning for image classification.'' in \emph{AAAI
  Conference on Artificial Intelligence (AAAI)}, 2011.

\bibitem{Lu_2017_TNNLS}
Y.~Lu, L.~Chen, A.~Saidi, E.~Dellandrea, and Y.~Wang, ``Discriminative transfer
  learning using similarities and dissimilarities,'' \emph{IEEE Transactions on
  Neural Networks and Learning Systems}, vol.~PP, no.~99, pp. 1--14, 2017.

\bibitem{krishna-cvpr2016}
K.~K. Singh, F.~Xiao, and Y.~J. Lee, ``Track and transfer: Watching videos to
  simulate strong human supervision for weakly-supervised object detection,''
  in \emph{IEEE Conference on Computer Vision and Pattern Recognition (CVPR)},
  2016.

\bibitem{Frome2013}
A.~Frome, G.~Corrado, J.~Shlens, S.~Bengio, J.~Dean, M.~Ranzato, and
  T.~Mikolov, ``Devise: A deep visual-semantic embedding model,'' in
  \emph{Neural Information Processing Systems (NIPS)}, 2013.

\bibitem{Bengio07_sae}
Y.~Bengio, P.~Lamblin, D.~Popovici, and H.~Larochelle, ``Greedy layer-wise
  training of deep networks,'' in \emph{Neural Information Processing Systems
  ({NIPS})}, 2007.

\bibitem{Fellbaum:98:Wordnet}
C.~Fellbaum, Ed., \emph{WordNet: An Electronic Lexical Database}.\hskip 1em
  plus 0.5em minus 0.4em\relax Cambridge, MA: MIT Press, 1998.

\bibitem{Leacock:98}
C.~Leacock and M.~Chodorow, ``Combining local context and {W}ord{N}et
  similarity for word sense identification,'' in \emph{{W}ord{N}et: An
  Electronic Lexical Database}, C.~Fellbaum, Ed.\hskip 1em plus 0.5em minus
  0.4em\relax MIT Press, 1998, pp. 265--283.

\bibitem{Resnik:95}
P.~Resnik, ``Using information content to evaluate semantic similarity in a
  taxonomy,'' in \emph{Proceedings of the International Joint Conference for
  Artificial Intelligence (IJCAI-95)}, 1995.

\bibitem{Lin:98}
D.~Lin, ``An information-theoretic definition of similarity,'' in
  \emph{International Conference on Machine Learning (ICML)}, 1998.

\bibitem{Mikolov:13}
T.~Mikolov, I.~Sutskever, K.~Chen, G.~Corrado, and J.~Dean, ``Distributed
  representations of words and phrases and their compositionality,'' in
  \emph{Neural Information Processing Systems ({NIPS})}, 2013.

\bibitem{Pennington:14}
J.~Pennington, R.~Socher, and C.~D. Manning, ``Glove: Global vectors for word
  representation,'' in \emph{Conference on Empirical Methods in Natural
  Language Processing (EMNLP)}, 2014, pp. 1532--1543.

\bibitem{Mikolov:13:NAACL}
T.~Mikolov, W.-t. Yih, and G.~Zweig, ``Linguistic regularities in continuous
  space word representations,'' in \emph{Conference of the North American
  Chapter of the Association for Computational Linguistics: Human Language
  Technologies (NAACL-HLT)}, 2013.

\bibitem{MisraExemplarSelection}
I.~Misra, A.~Shrivastava, and M.~Hebert, ``Watch and learn: Semi-supervised
  learning of object detectors from videos,'' in \emph{IEEE Conference on
  Computer Vision and Pattern Recognition (CVPR)}, 2015.

\bibitem{Rosenberg_WACV05}
C.~Rosenberg, M.~Hebert, and H.~Schneiderman, ``Semi-supervised self-training
  of object detection models,'' in \emph{IEEE Workshops on Application of
  Computer Vision (WACV)}, 2005.

\bibitem{Yang_cvpr13}
Y.~Yang, G.~Shu, and M.~Shah, ``Semi-supervised learning of feature hierarchies
  for object detection in a video,'' in \emph{IEEE Conference on Computer
  Vision and Pattern Recognition (CVPR)}, 2013.

\bibitem{agrawal14analyzing}
P.~Agrawal, R.~Girshick, and J.~Malik, ``Analyzing the performance of
  multilayer neural networks for object recognition,'' in \emph{European
  Conference on Computer Vision ({ECCV})}, 2014.

\bibitem{jia2014caffe}
Y.~Jia, E.~Shelhamer, J.~Donahue, S.~Karayev, J.~Long, R.~Girshick,
  S.~Guadarrama, and T.~Darrell, ``Caffe: Convolutional architecture for fast
  feature embedding,'' in \emph{ACM International Conference on Multimedia
  (MM)}, 2014.

\bibitem{ZeilerECCV2014}
M.~Zeiler and R.~Fergus, ``Visualizing and understanding convolutional
  networks,'' in \emph{European Conference on Computer Vision (ECCV)}, 2014.

\bibitem{rohrbach10cvpr}
M.~Rohrbach, M.~Stark, G.~Szarvas, I.~Gurevych, and B.~Schiele, ``What helps
  where {\textendash} and why? semantic relatedness for knowledge transfer,''
  in \emph{IEEE Conference on Computer Vision and Pattern Recognition
  ({CVPR})}, 2010.

\bibitem{Rothe:15}
S.~Rothe and H.~Sch\"{u}tze, ``Autoextend: Extending word embeddings to
  embeddings for synsets and lexemes,'' in \emph{the 53rd Annual Meeting of the
  Association for Computational Linguistics and the 7th International Joint
  Conference on Natural Language Processing (ACL)}, 2015, pp. 1793--1803.

\bibitem{gao_pcmp_cbmi12}
B.~Gao, E.~Dellandrea, and L.~Chen, ``Music sparse decomposition onto a midi
  dictionary of musical words and its application to music mood
  classification,'' in \emph{International Workshop on Content-Based Multimedia
  Indexing ({CBMI})}, 2012.

\bibitem{Simonyan14c}
K.~Simonyan and A.~Zisserman, ``Very deep convolutional networks for
  large-scale image recognition,'' in \emph{International Conference on
  Learning Representations (ICLR)}, 2015.

\bibitem{Szegedy_2016_CVPR}
C.~Szegedy, V.~Vanhoucke, S.~Ioffe, J.~Shlens, and Z.~Wojna, ``Rethinking the
  inception architecture for computer vision,'' in \emph{IEEE Conference on
  Computer Vision and Pattern Recognition (CVPR)}, 2016.

\bibitem{He_2016_CVPR}
K.~He, X.~Zhang, S.~Ren, and J.~Sun, ``Deep residual learning for image
  recognition,'' in \emph{IEEE Conference on Computer Vision and Pattern
  Recognition (CVPR)}, June 2016.

\bibitem{rbg_rcnn_pami}
R.~Girshick, J.~Donahue, T.~Darrell, and J.~Malik, ``Region-based convolutional
  networks for accurate object detection and segmentation,'' \emph{IEEE
  Transactions on Pattern Analysis and Machine Intelligence (TPAMI)}, vol.~38,
  no.~1, pp. 142--158, 2016.

\end{thebibliography}

\end{document}